\documentclass[runningheads]{llncs}

\usepackage{tikz}
\usepackage{comment}
\usepackage{graphicx}
\usepackage{amsmath}
\usepackage{amssymb}
\usepackage{booktabs}

\usepackage{times}
\usepackage{epsfig}
\usepackage{multirow}
\usepackage{times}
\usepackage{latexsym}
\usepackage[T1]{fontenc}
\usepackage[utf8]{inputenc}
\usepackage{microtype}
\usepackage{subcaption}
\usepackage{microtype}
\usepackage{enumitem}
\usepackage{rotating}
\usepackage{amsfonts}
\usepackage{amsmath}
\usepackage{soul}
\usepackage{mathtools}
\usepackage{chngcntr}
\usepackage{soul,xcolor}
\usepackage{pifont}
\usepackage{stackengine}
\usepackage{arydshln}
\usepackage[colorlinks,citecolor=green,urlcolor=black,bookmarks=false,hypertexnames=true]{hyperref}

\newcommand{\cmark}{\textcolor{green}{\ding{51}}}%
\newcommand{\xmark}{\textcolor{red}{\ding{55}}}%

\usepackage{cite}
\usepackage{wrapfig, lipsum, booktabs}

\newcommand{\modelrgb}{ViT-MT}
\newcommand{\modelseg}{TransLocator}

\newcommand{\ewa}[1]{\textcolor{red}{\emph{{Ewa:}~{#1}}}}

\begin{document}
\pagestyle{headings}
\mainmatter
\def\ECCVSubNumber{6867}  

\title{Where in the World is this Image? \\ Transformer-based Geo-localization in the Wild}

%
\author{Shraman Pramanick\inst{1},
Ewa M. Nowara\inst{1},
Joshua Gleason\inst{2},
\\ Carlos D. Castillo\inst{1}, and  Rama Chellappa\inst{1}}
\authorrunning{Pramanick et al.}
\institute{Johns Hopkins University \and
University of Maryland, College Park \\
\email{\{spraman3,carlosdc,rchella4\}@jhu.edu,\\ewa.m.nowara@gmail.com,gleason@umd.edu}}

\maketitle

\begin{abstract}

Predicting the geographic location (geo-localization) from a single ground-level RGB image taken anywhere in the world is a very challenging problem. The challenges include huge diversity of images due to different environmental scenarios, drastic variation in the appearance of the same location depending on the time of the day, weather, season, and more importantly, the prediction is made from a single image possibly having only a few geo-locating cues. For these reasons, most existing works are restricted to specific cities, imagery, or worldwide landmarks. In this work, we focus on developing an efficient solution to planet-scale single-image geo-localization. To this end, we propose \modelseg, a unified dual-branch transformer network that attends to tiny details over the entire image and produces robust feature representation under extreme appearance variations. TransLocator takes an RGB image and its semantic segmentation map as inputs, interacts between its two parallel branches after each transformer layer and simultaneously performs geo-localization and scene recognition in a multi-task fashion. We evaluate \modelseg\ on four benchmark datasets - Im$2$GPS, Im$2$GPS$3$k, YFCC$4$k, YFCC$26$k and obtain $5.5\%$, $14.1\%$, $4.9\%$, $9.9\%$ continent-level accuracy improvement over the state-of-the-art. \modelseg\ is also validated on real-world test images and found to be more effective than previous methods.


\end{abstract}

\section{Introduction}

\begin{figure}
\centering
\includegraphics[scale=0.5]{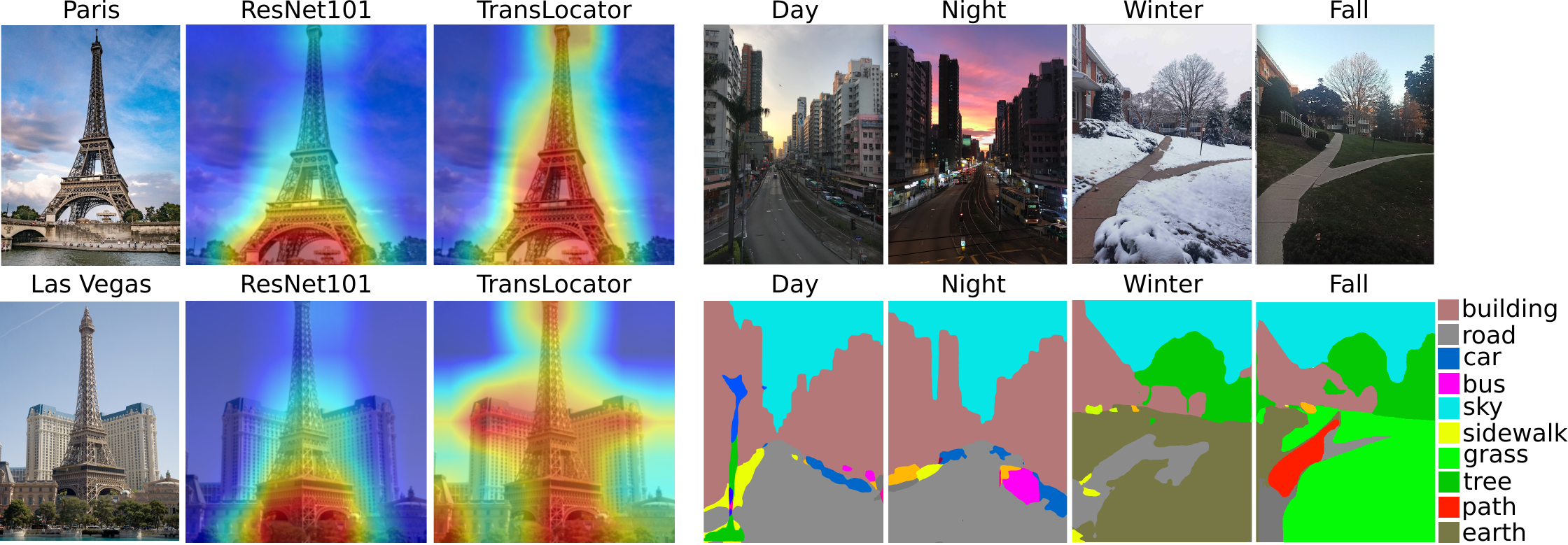}
\caption{\textit{First three columns: }\textbf{Utilizing global context - } \modelseg\ focuses on the entire image to correctly geo-locate the Eiffel Tower and its replica as visualized with Grad-CAM \cite{selvaraju2017grad} class activation maps. \textit{Last four columns: }\textbf{Robustness to appearance variations - }Semantic segmentation maps are robust to extreme appearance variations.}
    \label{fig:overview}
\end{figure}

Can we determine the location of a scene given a single ground-level RGB image? For famous and characteristic scenes, such as the Eiffel Tower, it is trivial because the landmark is so distinctive of Paris. Moreover, there is a lot of images captured in such prominent locations under different viewing angles, at different times of the day, and even in different weather or lighting conditions. However, some scenes, especially places outside cities and tourist attractions, may not have characteristic landmarks and it is not so obvious to locate where they were snapped. This is the case for the vast majority of the places in the world. Moreover, such places are less popular and are less photographed. As a result, there are very few images from such locations and the existing ones do not capture a diversity of viewing angles, time of the day, or weather conditions, making it much harder to geo-locate. Because of the complexity of this problem, most existing geo-localization approaches have been constrained to small parts of the world \cite{hausler2021patch, chen2011city, torii201524, seymour2018semantically}. Recently, convolutional neural networks (CNNs) trained with large datasets have significantly improved the performance of geo-localization methods and enabled extending the task to the scale of the entire world \cite{hays2008im2gps, hays2015large, weyand2016planet, muller2018geolocation}. However, planet-scale unconstrained geo-localization is still a very challenging problem and existing state-of-the-art methods struggle to geo-locate images taken anywhere in the world.

In contrast to many other vision applications, single-image geo-localization often depends on fine-grained visual cues present in small regions of an image. In Figure \ref{fig:overview}, consider the photo of the Eiffel Tower in Paris and its replica in Las Vegas. Even though these two images seem to come from the same location, the buildings and vegetation in the background play a decisive role in distinguishing them. Similarly, in the case of most other images, the global context spanned over the entire image is more important than individual foreground objects in geo-localization. Recently, a few studies \cite{doersch2020crosstransformers, raghu2021vision} comparing vision transformer (ViT) with CNNs have revealed that the early aggregation of global information using self-attention enables transformers to build long-range dependencies within an image. Moreover, higher layers of ViT maintain spatial location information better than CNNs. Hence, we argue that transformer-based networks are more effective than CNNs for geo-localization because they focus on detailed visual cues present in the entire image.

Another challenge of single-image geo-localization is the drastic appearance variation of the exact location under different daytime or weather conditions. Semantic segmentation offers a solution to this problem by generating robust representations in such extreme variations \cite{seymour2018semantically}. For example, consider the drastic disparity of the RGB images of same locations in day and night or winter and fall in Figure \ref{fig:overview}. In contrast to the RGB images, the semantic segmentation maps remain almost unchanged. Furthermore, semantic segmentation provides auxiliary information about the objects present in the image. This additional information can be a valuable pre-processing step since it enables the model to learn which objects occur more frequently in which geographic locations. For example, as soon as the semantic map detects mountains, the model immediately eliminates all flat regions, thus, reducing the complexity of the problem. 

Planet-scale geo-localization deals with a diverse set of input images caused by different environmental settings (e.g., outdoors vs indoors), which entails different features to distinguish between them. For example, to geo-locate outdoor images, features such as the architecture of buildings or the type of vegetation are important. In contrast, for indoor images, the shape and style of furniture may be helpful. To address such variations, Muller et al. \cite{muller2018geolocation} proposed to train different networks for different environmental settings. Though such an approach produces good results, they are cost-prohibitive and are not generalizable to a higher number of environmental scenarios. In contrast, we propose a unified multi-task framework for simultaneous geo-localization and scene recognition applied to images from all environmental settings.    
  
This work addresses the challenges of planet-scale single-image geo-localization by designing a novel dual-branch transformer architecture, \modelseg. We treat the problem as a classification task \cite{muller2018geolocation, weyand2016planet} by subdividing the earth's surface into a high number of geo-cells and assigning each image to one geo-cell. \modelseg\ takes an RGB image and its corresponding semantic segmentation map as input, divides them into non-overlapping patches, flattens the patches, and feeds them into two parallel transformer encoder modules to simultaneously predict the geo-cell and recognize the environmental scene in a multi-task framework. The two parallel transformer branches interact after every layer, ensuring an efficient fusion strategy. The resulting features learned by \modelseg\ are robust under appearance variation and focus on tiny details over the entire image.

In summary, our contributions are three-fold. $(i)$ We propose \modelseg\ - a unified solution to planet-scale single-image geo-localization with a dual-branch transformer network. \modelseg\ is able to distinguish between similar images from different locations by precisely attending to tiny visual cues. $(ii)$ We propose a simple yet efficient fusion of two transformer branches, which helps \modelseg\ to learn robust features under extreme appearance variation. $(iii)$ We achieve state-of-the-art performance on four datasets with a significant improvement of $5.5\%$, $14.1\%$, $4.9\%$, $9.9\%$ continent-level geolocational accuracy on Im$2$GPS \cite{hays2008im2gps}, Im$2$GPS$3$k \cite{hays2015large}, YFCC$4$k \cite{vo2017revisiting}, and YFCC$26$k \cite{theiner2022interpretable}, respectively. We also qualitatively evaluate the effectiveness of the proposed method on real-world images. Our source code is available at \url{https://github.com/ShramanPramanick/Transformer_Based_Geo-localization}.

\section{Related Works}


\subsection{Single-image geo-localization}

\noindent \textbf{Small-scale approaches:} Planet-scale single-image geo-localization is difficult due to several challenges, including the large variety of images due to different environmental scenarios and drastic differences in the appearance of same location based on the weather, time of day, or season. For this reason, many existing approaches are limited to geo-locating images captured in very specific and constrained locations. For example, many approaches have been restricted to geo-locating an image within a single city \cite{seymour2018semantically,hausler2021patch, berton2022rethinking}, such as Pittsburgh \cite{torii2013visual}, San Francisco \cite{chen2011city}, or Tokyo \cite{torii201524}. Far fewer approaches have attempted geo-localization in natural, non-urban environments. Some have limited the problem to only very specific natural environments, such as beaches~\cite{cao2012bluefinder, wang2013discovering}, deserts~\cite{tzeng2013user}, or mountains~\cite{baatz2012large,saurer2016image}.


\noindent \textbf{Cross-view approaches:} The challenge of large-scale image geo-localization with few landmarks and limited training data has led some researchers to propose cross-view approaches, which match a query ground-level RGB image with a large reference dataset of aerial or satellite images \cite{lin2013cross, tian2017cross, hu2018cvm, regmi2019bridging, toker2021coming, Zhu_2021_CVPR, yang2021cross, zhu2022transgeo}. However, these approaches require access to a relevant reference dataset which is not available for many locations.

\noindent \textbf{Planet-scale approaches:} Only a few approaches have attempted to geo-locate images on a scale of an entire world without any restrictions. Im$2$GPS \cite{hays2008im2gps} was the first work to geo-locate images taken anywhere on the earth using hand-crafted features and nearest neighbor search. The same authors have later improved their approach \cite{hays2015large} by refining the search with multi-class support vector machines. PlaNet~\cite{weyand2016planet} was the first deep learning approach for unconstrained geo-localization which significantly outperformed the two Im$2$GPS approaches \cite{hays2008im2gps,hays2015large}. Vo et al. \cite{vo2017revisiting} combined the Im$2$GPS and PlaNet approaches by using the deep-learning-based features to match a query image with a nearest neighbors search. The approach by Muller et al. \cite{muller2018geolocation} achieved the current state-of-the-art performance for unconstrained geo-localization. They fine-tuned three separate ResNet$101$ networks, each trained only on images of outdoor natural, outdoor urban, or indoor scenes. Detailed surveys of existing work on geo-localization can be found at \cite{brejcha2017state, masone2021survey}. 

\subsection{Vision Transformer}

Following the success of transformers \cite{vaswani2017attention} in machine translation, convolution-free networks that rely only on self-attentive transformer layers have gone viral in computer vision. In particular, Vision Transformer (ViT) \cite{dosovitskiy2020image} was the first to apply a pure transformer architecture on non-overlapping image patches and surpassed CNNs for image classification. Inspired by ViT, transformers have been widely adopted for various computer vision tasks, such as object detection \cite{carion2020end, zhu2020deformable, fang2021you, li2022grounded}, image segmentation \cite{xie2021segformer, zheng2021rethinking, strudel2021segmenter, cheng2022masked}, video understanding \cite{yang2019video, sun2019videobert, wang2021end, xu2021videoclip}, low-shot learning \cite{doersch2020crosstransformers, ye2020few}, image super resolution \cite{yang2020learning, chen2021pre},  $3$D classification \cite{zhao2021point, misra2021end}, and multimodal learning \cite{lu2019vilbert, li2020oscar, kant2020spatially, lei2021less, akbari2021vatt, hu2021unit, yang2022unified}. 

We are the first to address the ill-posed planet-scale single-image geo-localization problem by designing a novel dual-branch transformer architecture. Unlike \cite{muller2018geolocation}, we use only a single network for all kinds of input images, allowing our model to take advantage of the similar features from different scenes while maintaining its ability to learn scene-specific features. Moreover, our approach is robust under extreme appearance variations and generalizes to challenging real-world images.

\section{Proposed System - \modelseg}
This section presents our proposed multimodal multi-task system, \modelseg\, for geo-location estimation and scene recognition. Following the literature \cite{weyand2016planet, seo2018cplanet, muller2018geolocation}, we treat geo-localization as a classification problem by subdividing the earth into several geographical cells\footnote{Following \cite{muller2018geolocation}, we use the \textit{S$2$ geometry library} to generate the geo-cells.} containing a similar number of images. As the size and the number of geo-cells poses a trade-off between system performance and classification difficulty, we use multiple partitioning schemes which allow the system to learn geographical features at different scales, leading to a more discriminative classifier. Furthermore, we incorporate semantic representations of RGB images to learn robust feature representation across different daytime and weather conditions. Finally, we predict the geo-cell and environmental scenario of the input image by exploiting contextual information in a multi-task fashion.


\begin{wrapfigure}{L}{0.50\textwidth}

\vspace{-0.75 cm}
   \includegraphics[scale = 0.08]{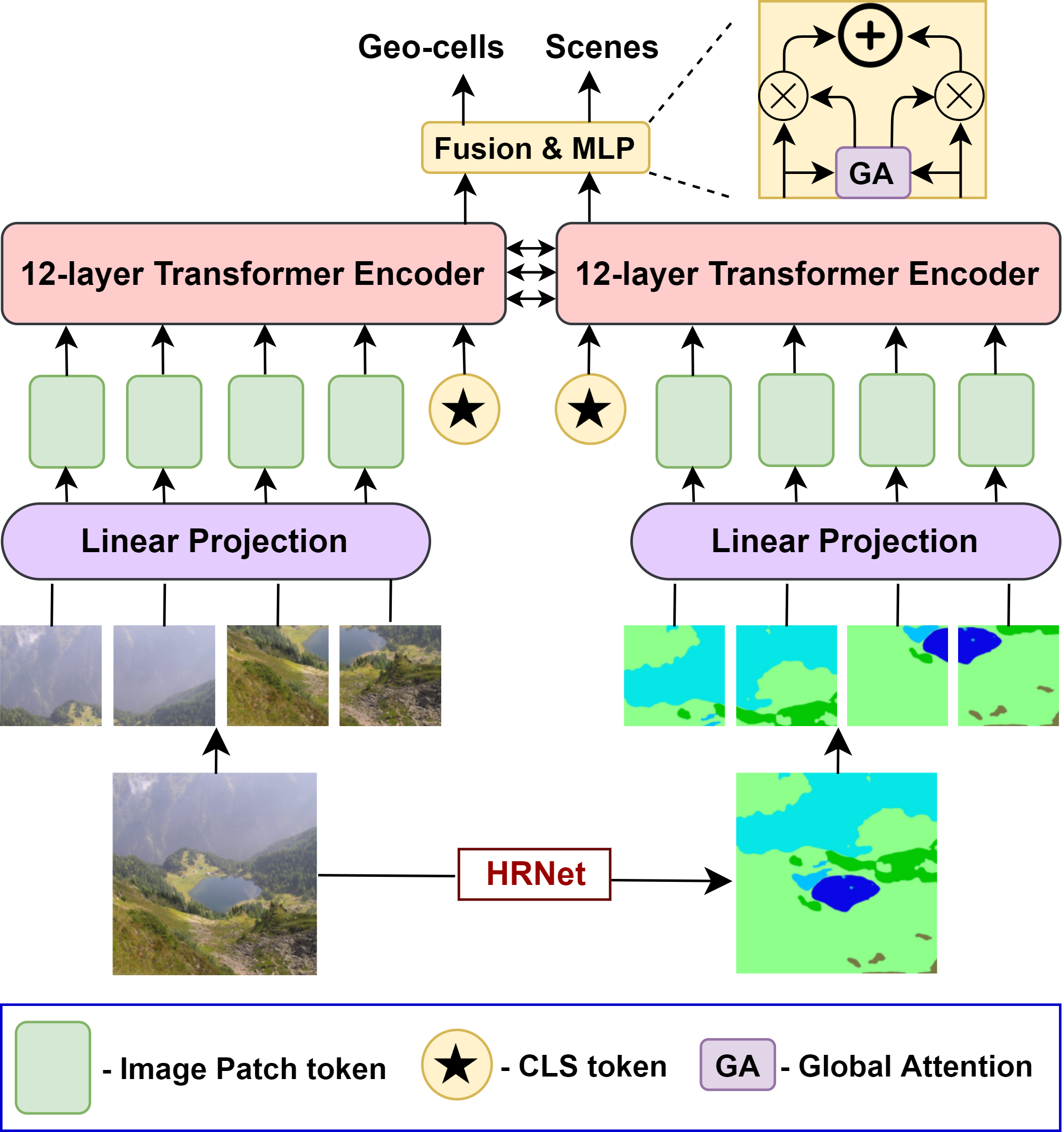}
   \caption{\textbf{Overview of the proposed model - \modelseg.} A dual-branch transformer network is trained using RGB images and corresponding semantic segmentation maps in a multi-task learning framework to simultaneously predict the environmental scene and the geographic location (geo-cell) of the image.}  
\label{fig:geolocalization_system}
\vspace{-0.4 cm}
\end{wrapfigure}

\subsection{Global Context with Vision Transformer} \label{sec:global_context}

We use a vision transformer (ViT) \cite{dosovitskiy2020image} model as the backbone of our architecture. Transformers have traditionally been used for sequential data, such as text or audio~\cite{vaswani2017attention, peters2018deep, devlin-etal-2019-bert, yang2019xlnet, liu2019roberta}. In order to extend transformers to vision problems, an image is split into a sequence of $2$-D patches which are then flattened and fed into the stacked transformer encoders through a trainable linear projection layer. An additional classification token (\texttt{CLS}) is added to the sequence, as in the original BERT approach~\cite{devlin-etal-2019-bert}. Moreover, positional embeddings are added to each token to preserve the order of the patches in the original image. A transformer encoder is made up with a sequence of blocks containing multi-headed self-attention (\texttt{MSA}) with a feed-forward network (\texttt{FFN}). \texttt{FFN} contains two multi-layer perceptron (\texttt{MLP}) layers with GELU non-linearity applied after the first layer. Layer normalization (\texttt{LN}) is applied before every block, and residual connections after every block. The output of $k$-th block  $x^{(k)}$ in the transformer encoder can be expressed as
\vspace{-0.1cm}
\begin{equation}
\vspace{-0.3cm}
    x^{(0)} = [x^{(0)}_{cls} || x^{(0)}_{patch}] + x_{pos} 
\end{equation}
\begin{equation}
\vspace{-0.15cm}
    y^{(k)} = x^{(k-1)} + \texttt{MSA}(\texttt{LN}(x^{(k-1)}))
\end{equation}
\begin{equation}
\vspace{-0.15cm}
    x^{(k)} = y^{(k)} + \texttt{FFN}(\texttt{LN}(y^{(k)}))
\end{equation}
where $x^{(i)}_{cls} \in \mathbb{R}^{1\times C}$ is the \texttt{CLS} token, $x^{(i)}_{patch} \in \mathbb{R}^{N\times C}$ is the patch tokens at $i$-th layer, and $x_{pos} \in \mathbb{R}^{(1+N)\times C}$ is the positional embeddings. $N$ and $C$ is the patch-sequence length and embedding dimension, respectively. $||$ denotes concatenation.





The self-attention in ViT is a mechanism to aggregate information from all spatial locations and is structurally very different from the fixed-sized receptive field of CNNs. Every layer of ViT can access the whole image and, thus, learn the global structure more effectively. This particular attribute of ViT plays a vital role in our classification system. In agreement with Raghu et al. \cite{raghu2021vision}, we empirically establish how the global receptive field of ViT can help to attend to small but essential visual cues, which CNNs often neglect. We provide a detailed comparative evaluation of this phenomenon in Section \ref{sec:interpretability}.         

\subsection{Semantic Segmentation for Robustness to Appearance Variation}

In order to improve the network's ability to generalize to scenes captured in different daytime and weather conditions, we train a dual-branch vision transformer on the RGB images and their corresponding semantic segmentation maps. As shown in Figure \ref{fig:overview}, compared to RGB features, the semantic layout of a scene is generally more robust to drastic appearance variations. We use HRNet \cite{wang2020deep} pre-trained on ADE20K and scene parsing datasets \cite{zhou2017scene, zhou2019semantic} to obtain high resolution semantic maps. HRNet assigns each pixel in the RGB image to one of the pre-defined $150$ object classes, such as \textit{sky}, \textit{tree}, \textit{sea}, \textit{building}, \textit{grass}, \textit{pier}, etc.  


\vspace{0.25cm}

\noindent \textbf{Multimodal Feature Fusion (MFF):} Our proposed framework contains two parallel transformer branches, one for the RGB image and the other for the corresponding semantic map. Since the two branches carry complementary information of the same input, effective fusion is the key for learning multimodal feature representations. We propose a simple and computationally light fusion scheme, where we sum the \texttt{CLS} tokens of each branch after every transformer encoder layer. At each layer, the \texttt{CLS} token is considered as an abstract global feature representation. This strategy is as effective as concatenating all feature tokens, but avoids quadratic complexity. Once the \texttt{CLS} tokens are fused, the information will be passed back to patch tokens at the later transformer encoder layers. More formally, $^{(i)} x^{(k)}$, the token sequence at $k$-th layer of a branch $i$ can be expressed as 
\begin{equation}
    ^{(i)} x^{(k)} = \Big[ g (\sum_{j \in \{ \text{rgb, seg} \} }   f(^{(j)} x^{(k)}_{\text{cls}})) ||  ^{(i)} x^{(k)}_{patch} \Big]
\end{equation}
where $f(.)$ and $g(.)$ are the projection and back-projection functions used to align the dimensions.

Since the relative importance of the two branches depends upon the structure of the input image, we attentively fuse the \texttt{CLS} tokens from the last layer of each branch. Motivated by \cite{gu2018hybrid, pramanick2021momenta, pramanick2022multimodal}, we design our attention module with two major parts -- modality attention generation and weighted concatenation. In the first part, a sequence of dense layers followed by a softmax layer is used to generate the attention scores $w_{mm} = [w_{rgb}, w_{seg}]$ for the two branches. In the second part, the \texttt{CLS} tokens from the last transformer layer are weighted using their respective attention scores and concatenated together as follows 
\begin{equation}
    f_{rgb} = (1 + w_{rgb}) ^{rgb} x^{(last)}_{cls}
\end{equation}
\begin{equation}
    f_{seg} = (1 + w_{seg}) ^{seg} x^{(last)}_{cls}
\end{equation}
\begin{equation}
    f_{mm} = [f_{rgb} || f_{seg}]
\end{equation}

We use residual connections to improve the gradient flow. The final multimodal representation, $ f_{mm}$, is fed into fully-connected classifier heads.


\subsection{Single Model with Multi-Task Learning}

Different features are essential for various environmental settings, such as indoor and outdoor urban or natural scenes. Hence, geo-localization can benefit from contextual knowledge about the surroundings, and this information can reduce the complexity of the data space, thus simplifying the classification problem. Muller et al. \cite{muller2018geolocation} addressed this issue and approached the problem by fine-tuning three individual networks separately on natural, urban, and indoor images. However, one immediate drawback of their approach is that using multiple separate networks is cost-prohibitive and limits the number of scene kinds. In addition, the separately trained networks can not share the learned features, which likely have semantic similarities across different scene kinds, which effectively reduces the size and potency of the training set.

We address these two drawbacks by training a single network with a multi-task learning objective for simultaneous geo-localization and scene recognition. Following \cite{muller2018geolocation}, we use a ResNet-$152$ pre-trained on \textit{Places$2$} dataset\footnote{\textit{Places$2$ ResNet$152$ model: \url{https://github.com/CSAILVision/places365}}} \cite{zhou2017places} to label the training images with corresponding $365$ different scene categories. Additionally, based on the provided scene hierarchy\footnote{\url{http://places2.csail.mit.edu/download.html}}, we label each image with a coarser $16$ and $3$ different scene labels. In multi-task learning, adding complementary tasks has been proven to improve the results of the main task \cite{caruana1997multitask, ranjan2017hyperface, zhang2021survey}. We follow a multi-task learning approach known as \textit{hard parameter sharing} \cite{ruder2017overview} which shares the parameters of hidden layers for all tasks and uses task-specific classifier heads. This strategy reduces overfitting because learning the same weights for multiple tasks forces the model to learn a generalized representation useful for the different tasks. More specifically, we only add a classifier head on top of the fused multimodal features and train the system end-to-end for both tasks.


\subsection{Training Objective} \label{sec:training_objective}

Since there is a trade-off between the classification difficulty and the prediction accuracy caused by the number and size of the geo-cells, we use partitioning of the earth's surface at three different resolutions, which we call \textit{coarse}, \textit{middle} and \textit{fine} geo-cells\footnote{Details on geo-cell partitioning can be found in the supplementary material.}. We feed the final multimodal feature representation $f_{mm}$ into four parallel classifier heads for the final classification: three for geo-cell prediction and one for scene recognition. We use a cross-entropy loss for each head. Our overall loss function can be summarized as follows:
\begin{equation}
\mathcal{L}_{total} =  (1 - \alpha - \beta) \mathcal{L}_{geo}^{coarse} + \alpha \mathcal{L}_{geo}^{middle} + \beta \mathcal{L}_{geo}^{fine} + \gamma \mathcal{L}_{scene}
\end{equation}

\vspace{-0.1cm}

\section{Experiments}

We conduct extensive experiments to show the effectiveness of our proposed method. In this section, we describe the datasets, the evaluation metrics, the baseline methods, and the detailed experimental settings.

\vspace{-0.15cm}
\subsection{Datasets}

We use publicly available RGB image datasets with corresponding ground truth GPS (latitude, longitude) tags for training, validation, and testing. We trained our model on the MediaEval Placing Task $2016$ dataset (MP-$16$) \cite{larson2017benchmarking} containing $4.72$M geo-tagged images sourced from Flickr. Like Vo et al. \cite{vo2017revisiting}, during training we excluded images taken by the same authors in our validation or test sets. We validated and tested our model on two randomly sampled subsets of images from the Yahoo Flickr Creative Commons $100$ Million dataset (YFCC$100$M)~\cite{thomee2016yfcc100m}, referred to as YFCC$26$k \cite{theiner2022interpretable} and YFCC$4$k \cite{vo2017revisiting} containing $25,600$ and $4536$ images, respectively. Since the images of MP-$16$, YFCC$26$k and YFCC$4$k were sourced without any scene and user restrictions, these datasets contain images of landmarks and landscapes, but also ambiguous images with little to no geographical cues, such as photographs of food and portraits of people.

We have additionally tested our model on two smaller datasets commonly used for geo-localization -- Im$2$GPS \cite{hays2008im2gps} and Im$2$GPS$3$k \cite{hays2015large}. Im$2$GPS contains $237$ manually selected geo-localizable images. In contrast to the previous three datasets, Im$2$GPS is specially designed to evaluate geo-localization systems and contains images from popular landmarks and famous tourist locations. Im$2$GPS$3$k is an extended version of Im$2$GPS. Im$2$GPS$3$k contains $2997$ images with geo-tags. Unlike Im$2$GPS, this dataset was not manually filtered and hence, it is a slightly more challenging test compared to Im$2$GPS.

\vspace{-0.15cm}
\subsection{Baselines}
We compare our method to several existing geo-estimation methods, including \textbf{Im$2$GPS} \cite{hays2008im2gps}, \textbf{[L]kNN} \cite{vo2017revisiting}, \textbf{MvMF} \cite{izbicki2019exploiting}, \textbf{Planet} \cite{weyand2016planet}, \textbf{CPlanet} \cite{seo2018cplanet}, and \textbf{ISNs} \cite{muller2018geolocation}. ISNs reports the state-of-the-art results on Im$2$GPS and Im$2$GPS$3$k. More details about the baselines are provided in the supplementary materials. Since neither of the baselines reported their performance on all considered test sets, we re-implement the missing ones. We also provide a detailed ablation study of our method by removing one component at a time from \modelseg\ in Table \ref{tab:results_ablation}. 



\subsection{Evaluation Metrics}

We evaluate the performance of our approach using geolocational accuracy at multiple error levels, i.e. the percentage of images correctly localized within a predefined distance from the ground truth GPS coordinates \cite{weyand2016planet, seo2018cplanet, muller2018geolocation}. Formally, if the predicted and ground truth coordinates are $(lat_{pred}, lon_{pred})$ and $(lat_{gt}, lon_{gt})$, the geo-locational accuracy $a_r$ at scale $r$ (in km) is defined as follows for a set of $N$ samples:
\begin{equation}
    a_r \equiv \dfrac{1}{N} \sum_{i=1}^N u(GCD((lat_{pred}^{(i)}, lon_{pred}^{(i)}), (lat_{gt}^{(i)}, lon_{gt}^{(i)})) - r)
\end{equation}
where GCD is the great circle distance and $u(x)$ = $\begin{cases} 1, \text{if } x < 0 \\ 0, \text{otherwise} \end{cases}$is an indicator function whether the distance is smaller than the tolerated radius $r$. We report the results at $1$ km, $25$ km, $200$ km, $750$ km, and $2500$ km ranges from the ground truth, which correspond to the scale of the same street, city, region, country, and continent, respectively \cite{weyand2016planet}.

\begin{wrapfigure}{R}{0.5\textwidth}
\vspace{-0.75 cm}
  \includegraphics[scale = 0.2525]{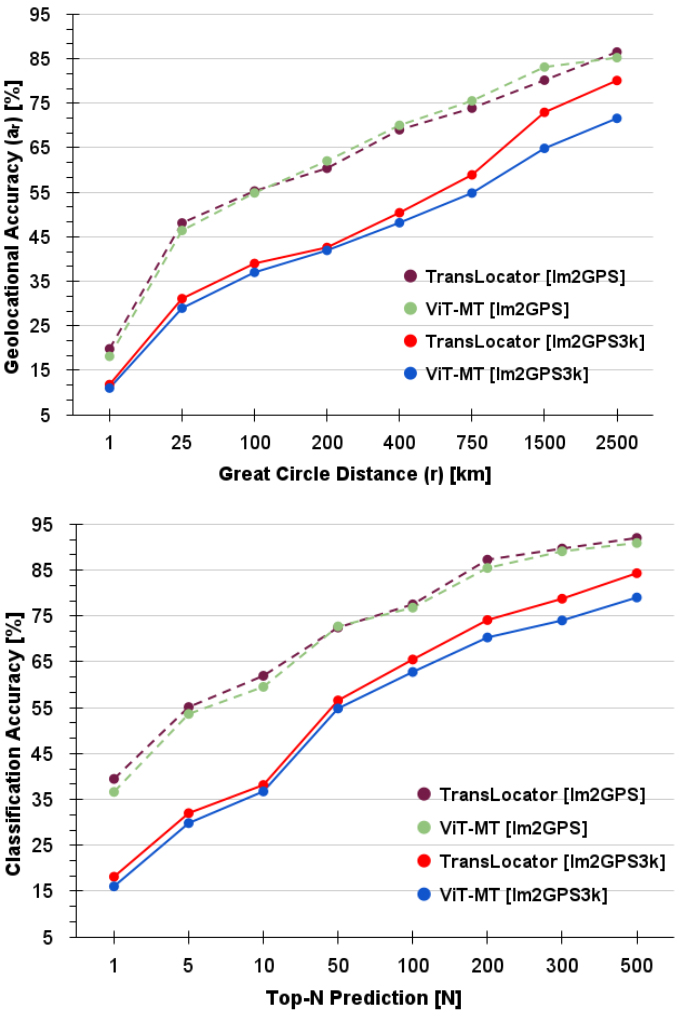}
  \caption{\textbf{Strong linear relationship between the geolocational and classification accuracy metrics.} The positive correlation enables us to treat geo-localization as a classification problem.}
\label{fig:metrices}
\vspace{-1 cm}
\end{wrapfigure}

Following Muller et al. \cite{muller2018geolocation}, we assign the predicted geo-cell a GPS tag by using the mean locations of all training images in that cell. Since the models are trained in a classification framework and evaluated using a distance metric, we empirically verified a strong linear relationship between classification and geo-locational accuracy. The average Pearson correlation coefficient between these two metrics on Im$2$GPS and Im$2$GPS$3$k test sets for \modelseg\ is $0.981$, which implies the validity of the classification framework for geo-localization (see Figure \ref{fig:metrices}). More details regarding the strong correlation between these two metrics can be found in the supplementary material.

\vspace{-0.1cm}
\subsection{Implementation Details}

We use ViT-B/$16$ \cite{dosovitskiy2020image} as the backbone of both RGB and segmentation channels. While training a single-channel (RGB only) multi-task transformer network, which we denote as \modelrgb, we use the pre-trained weights on the large ImageNet$21$K \cite{ridnik2021imagenet} dataset containing $14$ million images and $21$ thousand classes. We fine-tuned it for the geo-localization task. Both channels of the dual-branch \modelseg\ system are initialized with the weights of the \modelrgb\ backbone trained for $10$ epochs. The weights of the classifier heads are randomly initialized with
a zero-mean Gaussian distribution with a standard deviation of $0.02$. Following the standard ViT literature, we linearly project the non-overlapping patches of $16$ $\times$ $16$ pixels for both channels, and we add the \texttt{CLS} token and the positional embeddings. 

Since the training set contains images in various resolutions and scales, we use extensive data augmentation, as detailed in the supplementary materials. We implement the methods in Pytorch \cite{paszke2019pytorch} framework. Our \modelrgb\ and \modelseg\ took $6$/$10$ days to train on $20$ NVIDIA RTX 3090 GPUs, respectively, with $24$ GB dedicated memory in each GPU. We train both systems using a AdamW \cite{loshchilov2018decoupled} optimizer with an base learning rate of $0.1$, a momentum of $0.9$, and a weight decay of $0.0001$. We train the network for a total of $40$ epochs with a batch size of $256$. During testing, we convert the \textit{fine} geo-cells to corresponding GPS coordinates. Other necessary hyper-parameters and data augmentation details are given in the supplementary materials.   


\begin{table}[t!]
\centering 
\caption{\textbf{Geolocational accuracy of the proposed systems compared to several baselines across four datasets and five scales.} The methods re-implemented by~\cite{seo2018cplanet} are denoted by dagger($\dagger$), and the methods re-implemented by us are denoted by double-dagger($\ddagger$). \textcolor{blue}{$\Delta$} indicates improvement over state-of-the-art achieved by \modelseg.} 

\resizebox{0.8\columnwidth}{!}
{
\begin{tabular}{c | c || c c c c c}
\hline
\multirow{3}{*}{\bf Dataset} & \multirow{3}{*}{\bf \centering Method} & \multicolumn{5}{c}{\bf Distance ($a_r$ [\%] @ km)} \\ 

& & \multirow{1}{1.4 cm}{\tt \bf \centering Street} & \multirow{1}{1.4 cm}{\tt \bf \centering City} & \multirow{1}{1.4 cm}{\tt \bf \centering Region} & \multirow{1}{1.4 cm}{\tt \bf \centering Country} & \multirow{1}{1.4 cm}{\tt \bf \centering Continent} \\ 

& & \bf $1$ km & \bf $25$ km & \bf $200$ km & \bf $750$ km & \bf $2500$ km \\

\hline
\hline

\multirow{10}{1.2 cm}{\tt \bf \centering Im$2$GPS\\~\cite{hays2015large}} & \centering Human \cite{vo2017revisiting} & $-$ & $-$ & 3.8 & 13.9 & 39.3 \\



& \centering [L]kNN, $\sigma$ = $4$ \cite{vo2017revisiting} & 14.4 & 33.3 & 47.7 & 61.6 & 73.4 \\

& \centering MvMF \cite{izbicki2019exploiting} & 8.4 & 32.6 & 39.4 & 57.2 & 80.2\\

& \centering PlaNet \cite{weyand2016planet}  & 8.4 & 24.5 & 37.6 & 53.6 & 71.3\\

& \centering CPlaNet \cite{seo2018cplanet}  & 16.5 & 37.1 & 46.4 & 62.0 & 78.5\\

& \centering ISNs (M, f, S$_3$) \cite{muller2018geolocation} & 16.5 & 42.2 & 51.9 & 66.2 & 81.0 \\

& \centering ISNs (M,f$^*$,S$_3$) \cite{muller2018geolocation} & 16.9 & 43.0 & 51.9 & 66.7 & 80.2 \\  \cdashline{2-7}

& \modelrgb &  18.2 & 46.4 & 62.1 & 74.5 & 85.2 \\

& \modelseg & \bf 19.9 & \bf 48.1 & \bf 64.6 & \bf 75.6 & \bf 86.7 \\

\cline{2-7}

& \multicolumn{1}{c||}{\bf {\textcolor{blue}{$\Delta_{\texttt{Ours - ISNs}}$}}}  & \textcolor{blue}{3.0} \textcolor{blue}{$\uparrow$} & \textcolor{blue}{5.1} \textcolor{blue}{$\uparrow$} & \textcolor{blue}{12.7} \textcolor{blue}{$\uparrow$} & \textcolor{blue}{8.9} \textcolor{blue}{$\uparrow$} & \textcolor{blue}{5.5} \textcolor{blue}{$\uparrow$} \\

\hline
\hline

\multirow{8}{1.2 cm}{\tt \centering \bf Im$2$GPS\\$3$k\\ \cite{hays2008im2gps}} & [L]kNN, $\sigma$ = $4$ \cite{vo2017revisiting} & 7.2 & 19.4 & 26.9 & 38.9 & 55.9 \\

& PlaNet$^\dagger$ \cite{weyand2016planet} & 8.5 & 24.8 & 34.3 & 48.4 & 64.6 \\

& CPlaNet \cite{seo2018cplanet} & 10.2 & 26.5 & 34.6 & 48.6 & 64.6 \\

& ISNs (M, f, S$_3$) \cite{muller2018geolocation} & 10.1 & 27.2 & 36.2 & 49.3 & 65.6 \\

& ISNs (M,f$^*$,S$_3$) \cite{muller2018geolocation} & 10.5 & 28.0 & 36.6 & 49.7 & 66.0 \\  \cdashline{2-7}

& \modelrgb &  11.0 & 29.0 & 42.6 & 54.8 & 71.6 \\

& \modelseg &  \bf 11.8 & \bf 31.1 & \bf 46.7 & \bf 58.9 & \bf 80.1 \\

\cline{2-7}

& \multicolumn{1}{c||}{\bf {\textcolor{blue}{$\Delta_{\texttt{Ours - ISNs}}$}}} & \textcolor{blue}{1.3} \textcolor{blue}{$\uparrow$} & \textcolor{blue}{3.1} \textcolor{blue}{$\uparrow$} & \textcolor{blue}{6.1} \textcolor{blue}{$\uparrow$} & \textcolor{blue}{9.2} \textcolor{blue}{$\uparrow$} & \textcolor{blue}{14.1} \textcolor{blue}{$\uparrow$} \\

\hline 
\hline

\multirow{9}{1.2 cm}{\tt \centering \bf YFCC\\$4k$\\ \cite{theiner2022interpretable}} & [L]kNN, $\sigma$ = $4$ \cite{vo2017revisiting} & 2.3 & 5.7 & 11.0 & 23.5 & 42.0 \\

& PlaNet$^\dagger$ \cite{weyand2016planet} & 5.6 & 14.3 & 22.2 & 36.4 & 55.8 \\

& CPlaNet \cite{seo2018cplanet} & 7.9 & 14.8 & 21.9 & 36.4 & 55.5\\

& ISNs (M, f, S$_3$)$^\ddagger$ \cite{muller2018geolocation} & 6.5 & 16.2 & 23.8 & 37.4 & 55.0 \\

& ISNs (M,f$^*$,S$_3$)$^\ddagger$ \cite{muller2018geolocation} & 6.7 & 16.5 & 24.2 & 37.5 & 54.9 \\  \cdashline{2-7}

& \modelrgb & 8.1 & 18.0 & 26.2 & 40.0 & 59.9 \\

& \modelseg & \bf 8.4 & \bf 18.6 & \bf 27.0 & \bf 41.1  & \bf 60.4 \\

\cline{2-7}

& \multicolumn{1}{c||}{\bf {\textcolor{blue}{$\Delta_{\texttt{Ours - CPlanet}}$}}} & \textcolor{blue}{0.5} \textcolor{blue}{$\uparrow$} & \textcolor{blue}{3.8} \textcolor{blue}{$\uparrow$} & \textcolor{blue}{5.1} \textcolor{blue}{$\uparrow$} & \textcolor{blue}{4.7} \textcolor{blue}{$\uparrow$} & \textcolor{blue}{4.9} \textcolor{blue}{$\uparrow$} \\

& \multicolumn{1}{c||}{\bf {\textcolor{blue}{$\Delta_{\texttt{Ours - ISNs}}$}}} & \textcolor{blue}{1.7} \textcolor{blue}{$\uparrow$} & \textcolor{blue}{2.1} \textcolor{blue}{$\uparrow$} & \textcolor{blue}{2.8} \textcolor{blue}{$\uparrow$} & \textcolor{blue}{3.6} \textcolor{blue}{$\uparrow$} & \textcolor{blue}{5.5} \textcolor{blue}{$\uparrow$} \\

\hline 
\hline

\multirow{6}{1.2 cm}{\tt \centering \bf YFCC\\$26$k\\~\cite{vo2017revisiting}}& PlaNet$^\ddagger$ \cite{weyand2016planet} & 4.4 & 11.0 & 16.9 & 28.5 & 47.7\\

& ISNs (M, f, S$_3$)$^\ddagger$ \cite{muller2018geolocation} & 5.3 & 12.1 & 18.8 & 31.8 & 50.6 \\

& ISNs (M, f$^*$, S$_3$)$^\ddagger$ \cite{muller2018geolocation} &  5.3 & 12.3 & 19.0 & 31.9 & 50.7 \\ \cdashline{2-7}

& \modelrgb &  6.9 & 17.3 & 27.5 & 40.5 & 59.5 \\

& \modelseg & \bf 7.2 & \bf 17.8 & \bf 28.0 & \bf 41.3 & \bf 60.6\\

\cline{2-7}

& \multicolumn{1}{c||}{\bf {\textcolor{blue}{$\Delta_{\texttt{Ours - ISNs}}$}}} & \textcolor{blue}{1.9} \textcolor{blue}{$\uparrow$} & \textcolor{blue}{5.5} \textcolor{blue}{$\uparrow$} & \textcolor{blue}{9.0} \textcolor{blue}{$\uparrow$} & \textcolor{blue}{9.4} \textcolor{blue}{$\uparrow$} & \textcolor{blue}{9.9} \textcolor{blue}{$\uparrow$} \\

\hline 
\hline

\end{tabular}}

\label{tab:results_main}
\vspace{-3mm}
\end{table}

\vspace{-2mm}
\section{Results, Discussions and Analysis}

In this section, we compare the performance of \modelseg\ system with different baselines, and conduct a detailed ablation study to demonstrate the importance of different components in our system. Furthermore, we visualize the interpretability of \modelseg\ using Grad-CAM \cite{selvaraju2017grad} and perform an error analysis.

\begin{table}[t!]
\centering 
\caption{\textbf{Ablation Study on the Im$2$GPS and Im$2$GPS$3$K datasets.} \textit{Seg} denotes the segmentation branch of \modelseg, MFF represents multimodal feature fusion, and \textit{Scene} denotes the multi-task learning framework.} 
\vspace{2mm}

\resizebox{0.725\columnwidth}{!}
{
\begin{tabular}{c | c || c c c c c}
\hline
\multirow{3}{*}{\bf Dataset} & \multirow{3}{*}{\bf \centering Method} & \multicolumn{5}{c}{\bf Distance ($a_r$ [\%] @ km)} \\ 

& & \multirow{1}{1.4 cm}{\tt \bf \centering Street} & \multirow{1}{1.4 cm}{\tt \bf \centering City} & \multirow{1}{1.4 cm}{\tt \bf \centering Region} & \multirow{1}{1.4 cm}{\tt \bf \centering Country} & \multirow{1}{1.4 cm}{\tt \bf \centering Continent} \\ 

& & \bf $1$ km & \bf $25$ km & \bf $200$ km & \bf $750$ km & \bf $2500$ km \\

\hline
\hline

\multirow{6}{1.2 cm}{\tt \bf \centering Im$2$GPS\\ \cite{hays2015large}} & ResNet101 & 14.3 & 41.4 & 51.9 & 64.1 & 78.9 \\

& EfficientNet-B4 & 15.4 & 42.7 & 52.8 & 64.8 & 79.5\\ \cline{2-7}

& ViT base & 16.9 & 43.4 & 54.5 & 67.8 & 80.7\\

& + Seg & 17.6 & 44.8 & 58.9 & 70.0 & 83.3 \\

& + Seg + MFF & 19.0 & 47.2 & 62.7 & 73.5 & 85.7 \\

& \; + Seg + MFF + Scene \; & \bf 19.9 & \bf 48.1 & \bf 64.6 & \bf 75.6 & \bf 86.7 \\

\hline
\hline

\multirow{6}{1.2 cm}{\tt \bf \centering Im$2$GPS\\$3$k \\ \cite{hays2015large}}  & ResNet101 & 9.0 & 25.1 & 32.8 & 46.1 & 63.5 \\

& EfficientNet-B4 & 9.2 & 26.8 & 32.7 & 47.0 & 63.9 \\ \cline{2-7}

& ViT base & 9.9 & 28.0 & 37.8 & 54.2 & 70.7 \\

& + Seg & 10.5 & 29.1 & 42.5 & 55.8 & 73.6 \\

& + Seg + MFF & 11.1 & 30.2 & 45.0 & 56.8 & 78.1 \\

& \; + Seg + MFF + Scene \; & \bf 11.8 & \bf 31.1 & \bf 46.7 & \bf 58.9 & \bf 80.1 \\

\hline 
\hline

\end{tabular}}

\label{tab:results_ablation}
\vspace{-0.5cm}
\end{table}
\subsection{Comparison with Baselines}

Table \ref{tab:results_main} presents the performance of our proposed \modelseg\ system and baseline methods on all four evaluation datasets. The reported baselines have a similar number of training images and geographic classes, and hence, we can directly compare the results of our system with them. Since the Im$2$GPS and Im$2$GPS$3$k dataset mainly contains images of landmarks and popular tourist locations, the systems yield high accuracy on these two datasets. On Im$2$GPS, even the earlier methods like PlaNet and MvMF surpass human performance considerably. CPlaNet, a combinational geo-partitioning approach, brings a substantial improvement of $8.1\%$ in street-level accuracy over PlaNet. A similar trend of results is seen in the case of Im$2$GPS$3$k. On this dataset, CPlaNet beats PlaNet by $1.7\%$ street-level accuracy. For other distance scales, the results improve proportionally. The Individual Scene Networks (ISNs) report the state-of-the-art result on both of these datasets, achieving $16.5\%$ and $10.1\%$ street-level accuracy on Im$2$GPS and Im$2$GPS$3$k, respectively. The ensemble of hierarchical classifications (denoted by f$^*$) improves their results. However, note that none of these methods use semantic maps in their framework. Thus, we first implement the single-branch multi-task \modelrgb\ model. Interestingly, even this model produces a significant improvement over ISNs for all scales on both datasets, which can be explained by the global context used by ViT architecture. Our final dual-branch \modelseg\ system improves on top of \modelrgb. Overall, we push the current state-of-the-art by \textbf{$3.0\%$} and \textbf{$1.3\%$} street-level and \textbf{$5.5\%$} and \textbf{$14.1\%$} continent-level accuracy on Im$2$GPS and Im$2$GPS$3$k, respectively.

The YFCC$4$k and YFCC$26$k datasets contain more challenging samples which have little to no geo-locating cues. However, these large datasets of unconstrained real-world images examine the generalizability of the systems. On YFCC$4$k, CPlaNet produces the best street-level accuracy among baselines, while ISNs beats CPlaNet at the continent level. Our proposed \modelseg\ outperforms both CPlaNet and ISNs in every distance scale, improving the state-of-the-art by \textbf{$1.7\%$} and \textbf{$5.5\%$} street-level and continent-level accuracy. The YFCC$26$k dataset is even more challenging than YFCC$4$k; the best baseline produces only $5.3\%$ street-level accuracy. However, our proposed \modelseg\ system yields an impressive $7.2\%$ street-level accuracy on YFCC$26$k, which proves the appreciable generalizability of \modelseg.


\begin{table} [t!]
\centering 
\caption{\textbf{Effect of number of scenes on \modelseg.} Fine-grained scene information helps \modelseg\ to achieve superior performance.}
\vspace{2mm}
\resizebox{0.725\columnwidth}{!}
{
\begin{tabular}{c | c | c || c c c c c}
\hline
\multirow{3}{*}{\bf Dataset} & \multirow{3}{*}{\bf Method} & \multirow{3}{*}{\bf \#Scenes} & \multicolumn{5}{c}{\bf Distance ($a_r$ [\%] @ km)} \\ 

& & & \multirow{1}{1.4 cm}{\tt \bf \centering Street} & \multirow{1}{1.4 cm}{\tt \bf \centering City} & \multirow{1}{1.4 cm}{\tt \bf \centering Region} & \multirow{1}{1.4 cm}{\tt \bf \centering Country} & \multirow{1}{1.4 cm}{\tt \bf \centering Continent} \\ 

& & & \bf $1$ km & \bf $25$ km & \bf $200$ km & \bf $750$ km & \bf $2500$ km \\
\hline
\hline

\multirow{4}{1.2 cm}{\tt \bf \centering Im$2$GPS \\ \cite{hays2008im2gps}} & \multirow{3}{*}{\; \modelseg\ \;} & 3 & 18.4 & 46.3 & 55.6 & 68.6 & 84.0 \\
& & 16 & 19.0 & 47.1 & 56.5 & 69.7 & 85.4\\
& & 365 & \bf 19.9 & \bf 48.1 & \bf 57.4 & \bf 70.9 & \bf 86.5 \\\cdashline{2-8}
& \multicolumn{2}{c||}{\textcolor{blue}{\bf \centering $\Delta_{\texttt{Scenes\textsubscript{365} - Scenes\textsubscript{3}}}$}} & \textcolor{blue}{1.5 $\uparrow$} & \textcolor{blue}{1.8 $\uparrow$} & \textcolor{blue}{1.8 $\uparrow$} & \textcolor{blue}{2.3 $\uparrow$} & \textcolor{blue}{2.5 $\uparrow$} \\ 

\hline
\hline

\multirow{4}{1.2 cm}{\tt \bf \centering Im$2$GPS\\$3$k \\ \cite{hays2015large}} & \multirow{3}{*}{\; \modelseg\ \;} & 3 & 10.8 & 29.9 & 41.0 & 56.8 & 78.7 \\
& & 16 & 11.6 & 30.5 & 42.1 & 57.6 & 79.4 \\
& & 365 & \bf 11.8 & \bf 31.1 & \bf 42.6 & \bf 58.9 & \bf 80.1\\\cdashline{2-8}
& \multicolumn{2}{c||}{\textcolor{blue}{\bf \centering $\Delta_{\texttt{Scenes\textsubscript{365} - Scenes\textsubscript{3}}}$}} & \textcolor{blue}{1.0 $\uparrow$} & \textcolor{blue}{1.2 $\uparrow$} & \textcolor{blue}{1.6 $\uparrow$} & \textcolor{blue}{2.1 $\uparrow$} & \textcolor{blue}{1.4 $\uparrow$} \\ 
\hline 
\hline

\end{tabular}}

\label{tab:scenes}
\end{table}

\subsection{Ablation Study} \label{sec:ablation_results}

\noindent \textbf{Role of Vision Transformer:} We conduct a detailed ablation study to understand the contributions of different components proposed in \modelseg\ architecture. We start by comparing a base ViT-B/16 encoder with two conv networks - ResNet101 \cite{he2016deep} and EfficientNet-B4 \cite{tan2019efficientnet}. We re-train these three systems using only RGB images of the MP-$16$ dataset. As shown in Table \ref{tab:results_ablation}, the base ViT architecture produces consistent improvements over the conv models on Im$2$GPS and Im$2$GPS$3$k, which confirms the effectiveness of the larger receptive field of ViT for geo-localization. 

\begin{figure}[hbtp]
\centering
\hspace*{0cm}
\begin{center}
  \includegraphics[scale=0.1075]{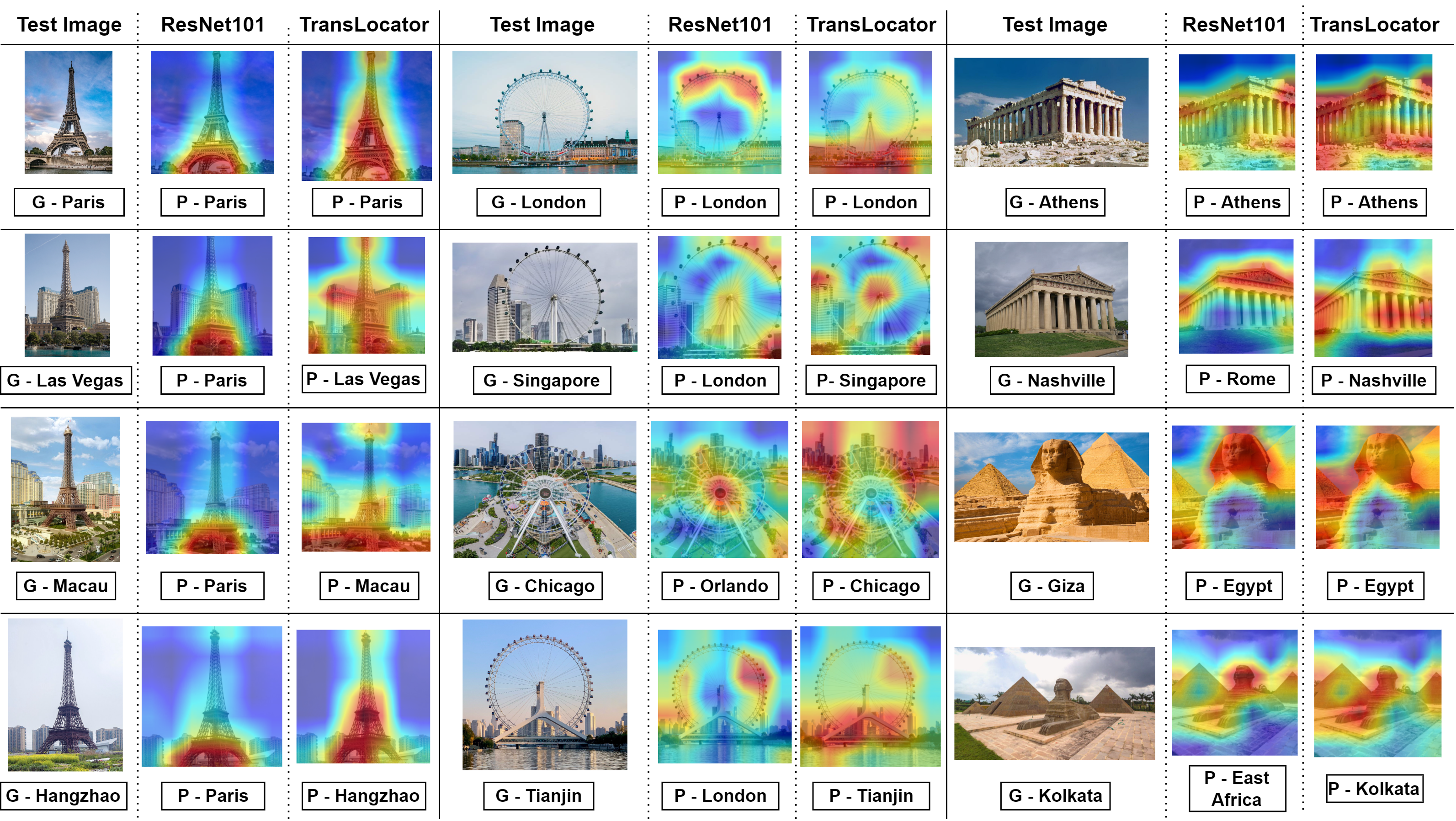}
\end{center}
\vspace{-0.4cm}
  \caption{\textbf{Qualitative comparison of \modelseg\ and ResNet$101$ on images with similar landmarks but from different geographic locations.} Unlike ResNet$101$, \modelseg\ focuses both on the foreground and background and is able to correctly geo-locate the very similar looking images. \textit{G} and \textit{P} denotes ground truth and predicted location.}
  \label{fig:gradcam}
\end{figure}

\noindent \textbf{Role of Segmentation Maps:} We then add the segmentation branch to base-ViT by concatenating the \texttt{CLS} tokens from the last layers. This method yields an improvement of $0.7\%$ over base-ViT on Im$2$GPS. However, this kind of fusion is not optimal because the two channels do not interact in-between. Hence, we then incorporate our proposed multimodal feature fusion (MFF), which sums \texttt{CLS} tokens after each transformer layer. MFF improves on the base-ViT by $2.1\%$ and $1.2\%$ street-level accuracy on Im$2$GPS and Im$2$GPS$3$k,  suggesting that the two branches learn complementary and robust features for different images.

\noindent \textbf{Role of Multi-task Learning:} Next, we incorporate an additional classifier head for scene recognition, which significantly improves the performance. Moreover, we evaluate the effect of coarse- and fine-grained contextual knowledge by varying the number of scene categories. As shown in Table \ref{tab:scenes}, a higher number of scenes improves street-level accuracy by $1 - 1.5\%$ across the two datasets. 

\subsection{Interpretability of \modelseg} \label{sec:interpretability}

In this section, we comprehend the interpretability of \modelseg\ by generating visual explanations using Grad-CAM \cite{selvaraju2017grad}. First, we focus on similar-looking images coming from different portions of the world. As shown in Figure \ref{fig:gradcam}, these images\footnote{Collected from the Internet under creative commons license.} can be discriminated only by close attention to the global context.

\begin{wrapfigure}{r}{0.50\textwidth}
\vspace{-3mm}
  \includegraphics[scale=0.132]{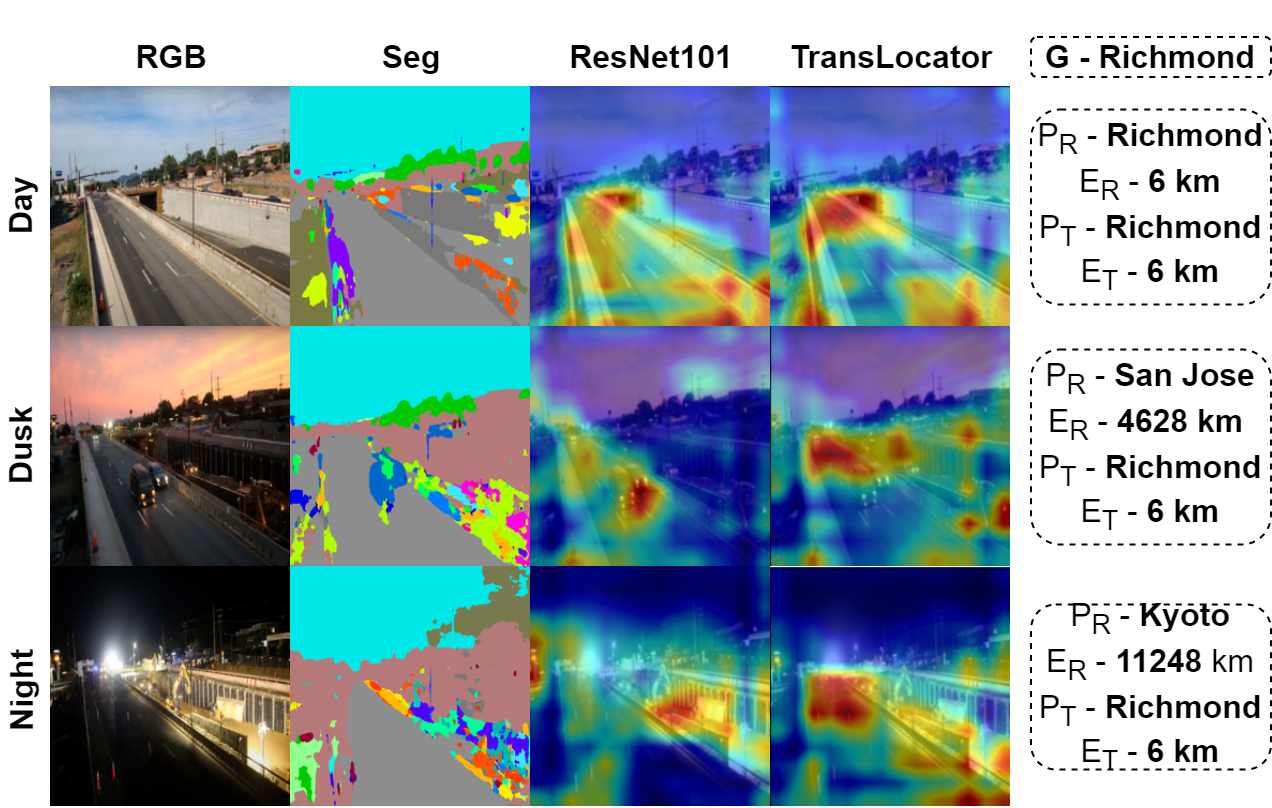}
  \caption{\textbf{Qualitative comparison of \modelseg\ and ResNet$101$ on images with same location but under challenging appearance variations.} Unlike ResNet$101$, \modelseg\ attends to similar regions in each image and locates all three images correctly. \textit{G} denotes ground truth, \textit{$P_R$}, \textit{$E_R$}, \textit{$P_T$} and \textit{$E_T$} denotes predicted location and prediction error by ResNet$101$ and \modelseg, respectively.}
 \label{fig:gradcam_daytime_variation}
\vspace{-6 mm}
\end{wrapfigure}

For example, the famous Eiffel Tower in Paris closely resembles the towers in Las Vegas, Macau, and Hangzhou. However, there are differences in the background pixels between these images, such as different characteristic buildings and vegetation. While the ResNet$101$ architecture fails to locate the image from Las Vegas, our proposed \modelseg\ network correctly discriminates it from the Eiffel Tower in Paris by adequately focusing on the background structure. The similar superior performance of \modelseg\ is also shown for other images containing Ferris Wheel, Parthenon-like, and pyramid-like constructions in different geographic locations.

Next, we investigate the case of drastic appearance variation in the same location. Figure \ref{fig:gradcam_daytime_variation} shows three images\footnote{Collected from \url{https://www.virginiadot.org} under creative commons license.} taken in Richmond, Virginia by a highway surveillance camera in the morning, dusk, and night. Though the RGB image varies with the change in the daytime, the corresponding semantic segmentation maps remain similar, and thus \modelseg\ can learn robust multimodal features. In contrast to ResNet$101$, which suffers from such appearance variation, \modelseg\ attends to similar regions in each image and locates all three images correctly.

\subsection{Error Analysis}

\begin{wrapfigure}{r}{0.50\textwidth}
\vspace{-0.9 cm}
   \includegraphics[scale = 0.1655]{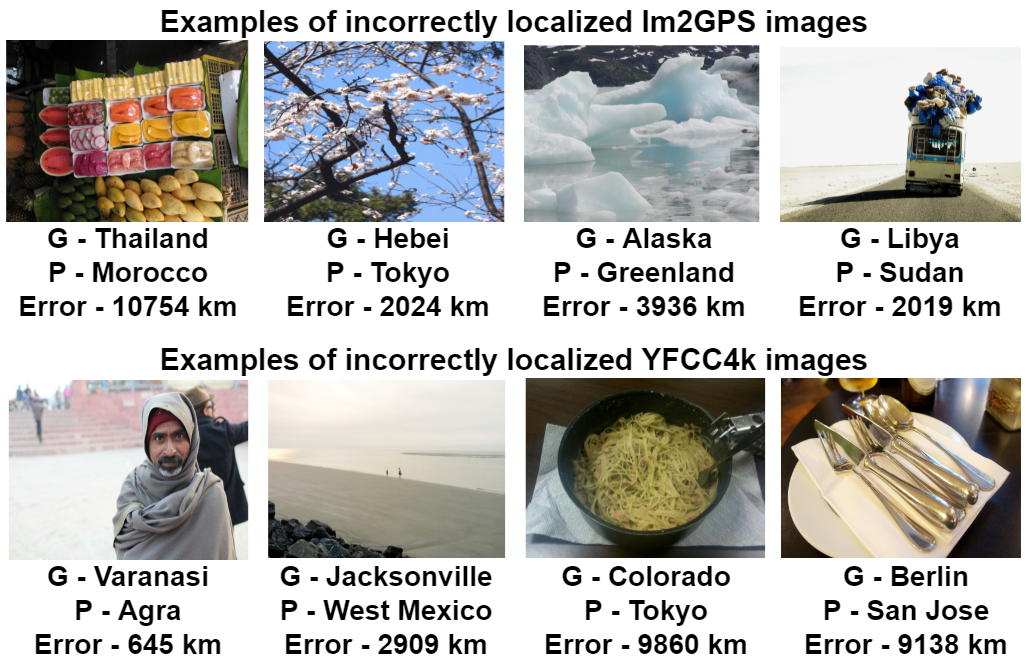}
   \caption{\textbf{Limitations of our method:} images without salient geo-locating cues cannot be geo-located correctly. \textit{G} and \textit{P} denotes ground truth and predicted location.}
\label{fig:limitations}
\end{wrapfigure}

Although \modelseg\ achieves better quantitative and qualitative results than baselines, there are still some open geo-localization problems that \modelseg\ does not solve. \modelseg\ can not locate images without geo-locating cues. For example, a photo of a cherry blossom tree can come anywhere from Tokyo, Paris, Washington DC, Dublin, etc. Hence \modelseg\ can not locate such an image. Similarly, pictures of sea beaches, deserts, or pictures of foods with no background cues can never precisely be geo-located. A few samples of such incorrectly located images are illustrated in Figure \ref{fig:limitations}.


\section{Conclusion}
Planet-scale single-image geo-localization is a highly challenging problem. These challenges include images with a large diversity in various environmental scenarios and appearance variation due to daytime, season, or weather changes. Hence, most existing approaches limit geo-localization in the scale of landmarks, a specific area, or an environmental scenario. Some approaches propose to use separate systems for different environments. In this paper, we address this challenging problem by proposing \modelseg, a unified dual-branch transformer network that attends to tiny details over the entire image and produces robust feature representation under extreme appearance variations. \modelseg\ takes an RGB image with its semantic segmentation map as input, interacts between its two parallel channels after each transformer layer, and concatenates the learned RGB and semantic representations using global attention. We train \modelseg\ in a unified multi-task framework for simultaneous geo-localization and scene recognition, and thus, our system can be applied to images from all environmental settings. Extensive experiments with \modelseg\ on four benchmark datasets - Im$2$GPS \cite{hays2008im2gps}, Im$2$GPS$3$k \cite{hays2015large}, YFCC$4$k \cite{vo2017revisiting} and YFCC$26$k \cite{theiner2022interpretable} shows a significant improvement of $5.5\%$, $14.1\%$, $4.9\%$, $9.9\%$ continent-level accuracy over current state-of-the-art. We also obtain better qualitative results when we test \modelseg\ on challenging real-world images. \\
\noindent\textbf{Acknowledgement:} This research is partially supported by an ARO MURI Grant No. W$911$NF-$17$-$1$-$0304$.    

\bibliographystyle{splncs04}
\bibliography{main}

\newpage
\thispagestyle{plain}
\makeatletter
{\Large \bf \centering Where in the World is this Image? \\ Transformer-based Geo-localization in the Wild \\ (Supplementary Material) \par \bigskip \bigskip \bigskip}
\appendix

\counterwithin{figure}{section}
\numberwithin{table}{section}

\vspace{-3.5mm}

In this supplementary material, we provide additional details on geo-cell partitioning, data augmentation, hyper-parameter values, baselines, evaluation metrics and illustrate additional quantitative and qualitative results. 

\section{Implementation Details $\&$ Hyper-parameter Values}
\subsection{Adaptive Geo-cell Partitioning}

We utilize the \textit{S$2$ geometry library}\footnote{\url{https://code.google.com/archive/p/s2-geometry-library/source}} to divide the earth's surface into a fixed number of non-overlapping geographic cells. To directly compare our results with baselines, we use the same partitioning approach like \cite{muller2018geolocation}, where we subdivide the earth's surface into three resolutions containing $3298$, $7202$, and $12893$ geo-cells referred to as coarse, middle, and fine cells, respectively. The partitioning ensures each cell contains at least $50$ and at most $5000$, $2000$, and $1000$ training images for the coarse, middle, and fine resolution. Limiting the number of training images into a minimum and maximum range per geo-cell gives two advantages. First, the training set does not suffer from class imbalance, which is pivotal for classifying many classes. Second, the geographic areas which are heavily photographed are subdivided into smaller cells, allowing more precise geo-localization of these regions (such as big cities and tourist attractions). However, one drawback of this approach is that many geographic areas have less than the required minimum number of images. Consequently, many locations (such as oceans, remote mountainous regions, deserts, poles) are discarded because of insufficient images. With the minimum range of $50$, the partitioning covers almost $84\%$ of the entire earth's surface.

\begin{figure}[t!]
\centering
\hspace*{0cm}
\begin{center}
  \includegraphics[scale=0.06]{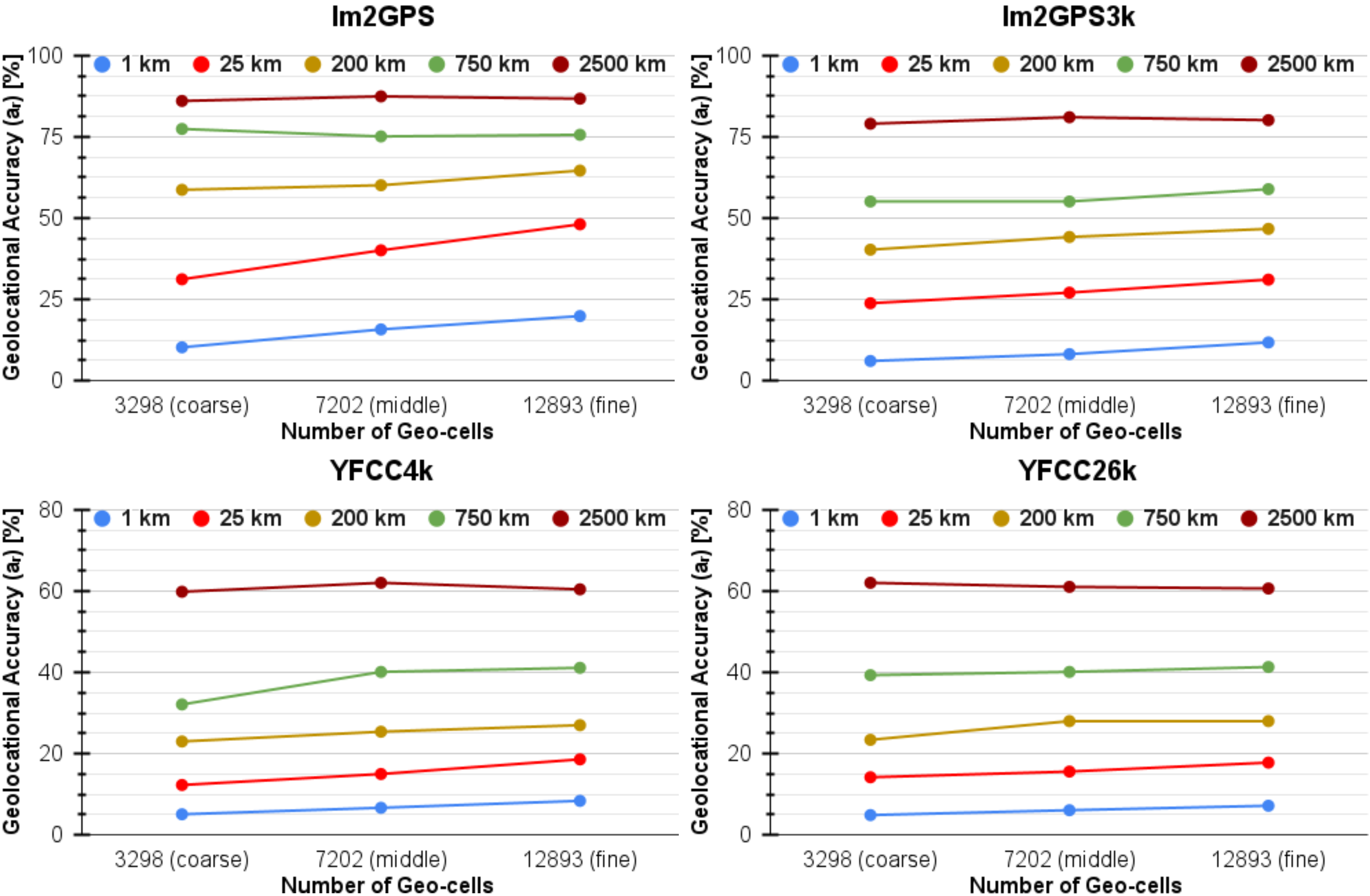}
\end{center}
\vspace{-0.5cm}
  \caption{\textbf{Effect of the predictions from three different partitioning schemes on the performance of \modelseg.} The street-level ($1$ km) and city-level ($25$ km) geolocational accuracy significantly improves as we employ finer geo-cells. However, the continent-level ($2500$ km) accuracy remains almost unchanged.}
  \label{fig:geo-cells_supplementary}
\vspace{-0.5cm}
\end{figure}

In classification-based geo-localization, the number of geo-cells closely relates to the prediction accuracy. In other words, since the predicted GPS coordinates are always the mean location of all training images in the predicted geo-cell, coarse cells often can not produce good street-level accuracy. On the other hand, fine cells improve localization precision by generating smaller geo-cells in highly photographed areas. Figure \ref{fig:geo-cells_supplementary} shows the improvement of street-level ($1$ km) and city-level ($25$ km) geolocational accuracy by using the predictions from finer cells on four different datasets. Since the continent-level ($2500$ km) accuracy is not directly related to the size of cells, it remains almost unchanged with geo-cell resolution variation. Next, we use an ensemble of hierarchical classification using all three resolutions. However, in agreement with \cite{theiner2022interpretable}, this method does not achieve a consistent improvement than considering only fine partitioning. Moreover, the ensemble increases inference time by almost $9\%$. Following these observations, we use the predictions of the fine geo-cells in all our experiments. We believe that adding a retrieval network after classification would improve the performance by allowing the system to search within the predicted geo-cell. However, this paper does not consider any retrieval extensions for a fair comparison with the baselines.     

\vspace{-0.3cm}
\subsection{Data Augmentation} \label{sec:data_augmentation}
\vspace{-0.1cm}

Since the training set contains images in various orientations, resolutions and scales, we use extensive data augmentation. The augmentation policy includes: \textit{RandomAffine} with degrees ($0$, $15$), \textit{ColorJitter} containing $\{$brightness, contrast, saturation, hue$\}$ strength of $\{$$0.4$, $0.4$, $0.4$, $0.1$$\}$ with a probability of $0.8$, \textit{RandomHorizontalFlip} with a probability of $0.5$, \textit{Resize} in ($256$, $256$), \textit{Tencrop} with size ($224$, $224$) and standard \textit{Normalization}. We apply \textit{ColorJitter} only in the RGB channel. Table \ref{tab:results_augmentation} shows an empirical analysis of the effectiveness of different augmentation techniques on training \modelseg. We start with the standard \textit{Flip}, \textit{Resize} and \textit{Normalization} operations. Adding \textit{Affine} transformation and \textit{ColorJitter} helps in improving the performance by a tiny margin. However, the \textit{TenCrop} augmentation shows to have a significant effect, improving $0.5-1.1\%$ street-level accuracy in Im$2$GPS and Im$2$GPS$3$k datasets. Since the important visual cues for geo-localization often reside on the edges of the image, taking multiple crops from different positions and averaging the predictions helps in improving the performance. 


\begin{table}[t!]
\centering 
\caption{\textbf{Role of different data augmentation techniques on training \modelseg.} RHF, R, N, RA, CJ and TC denotes \textit{RandomHorizontalFlip}, \textit{Resize}, \textit{Normalization}, \textit{RandomAffine}, \textit{ColorJitter} and, \textit{Tencrop}, respectively, using the parameters mentioned in Section \ref{sec:data_augmentation}.} 
\vspace{2mm}

\resizebox{0.775\columnwidth}{!}
{
\begin{tabular}{c | c || c c c c c}
\hline
\multirow{3}{*}{\bf Dataset} & \multirow{3}{*}{\bf \centering Method} & \multicolumn{5}{c}{\bf Distance ($a_r$ [\%] @ km)} \\ 

& & \multirow{1}{1.4 cm}{\tt \bf \centering Street} & \multirow{1}{1.4 cm}{\tt \bf \centering City} & \multirow{1}{1.4 cm}{\tt \bf \centering Region} & \multirow{1}{1.4 cm}{\tt \bf \centering Country} & \multirow{1}{1.4 cm}{\tt \bf \centering Continent} \\ 

& & \bf $1$ km & \bf $25$ km & \bf $200$ km & \bf $750$ km & \bf $2500$ km \\

\hline
\hline

\multirow{4}{1.2 cm}{\tt \bf \centering Im$2$GPS\\ \cite{hays2015large}} & \; RHF + R + N \; & 18.8 & 46.2 & 62.8 & 73.6 & 83.6  \\
& + RA & 18.8 & 46.5 & 63.1 & 73.8 & 83.8 \\
& + RA + CJ & 19.0 & 46.8 & 63.2 & 74.1 & 84.0 \\
& \; + RA + CJ + TC \; & \bf 19.9 & \bf 48.1 & \bf 64.6 & \bf 75.6 & \bf 86.7 \\

\hline
\hline

\multirow{4}{1.2 cm}{\tt \bf \centering Im$2$GPS\\$3$k\\ \cite{hays2015large}} & RHF + R + N & 11.3 & 30.4 & 45.7 & 58.0 & 78.4 \\
& + RA & 11.4 & 30.6 & 46.0 & 58.0 & 78.5 \\
& + RA + CJ & 11.4 & 30.8 & 45.9 & 58.2 & 78.7 \\
& + RA + CJ + TC & \bf 11.8 & \bf 31.1 & \bf 46.7 & \bf 58.9 & \bf 80.1 \\

\hline 
\hline

\end{tabular}}

\label{tab:results_augmentation}
\vspace{-0.5cm}
\end{table}

\begin{table}[t]
\centering 
\caption{\textbf{Hyper-parameters of \modelseg.}}
\vspace{2mm}
\resizebox{0.55\textwidth}{!}
{
\begin{tabular}{l|c|c}
\hline
\textbf{Hyper-parameters} & \textbf{Notation} & \textbf{Value} \\
\hline
\hline
\#dim for dense layers in MFF & - & $[768, 8, 1]$ \\ \cdashline{1-3}
\multirow{4}{*}{{\#dim} for classification FC} & \centering coarse & $[768, 3298]$ \\
& \centering middle & $[768, 3000, 7202]$ \\
& \centering fine & $[768, 6000, 12893]$  \\
& \centering scene & $[768, 3/16/365]$ \\
\hline
\multicolumn{3}{c}{Training} \\ \hline
{Batch-size} & - & $256$ \\
{Epochs} & $N$ & $40$ \\ 
{Optimizer} & - & AdamW \\ 
{Loss} & - & CE \\ 
{Base learning rate} & $\alpha$ & $0.1$ \\
Momentum & - & 0.9 \\
Learning rate scheduler & - & Cosine \\
Warmup epochs & - & $2$ \\
Weight decay & - & $0.0001$ \\

\hline
\end{tabular}}
\label{tab:hyperparameters}
\vspace{-0.5cm}
\end{table}

\vspace{-0.3cm}
\subsection{Hyper-parameter Details}

In Table \ref{tab:hyperparameters}, we furnish the details of hyper-parameters used during training. Grid search is performed on batch size, learning rate, and the depth of classifier heads to find the best hyper-parameter configuration. The model is evaluated after every epoch on the validation set and the best model was taken to be evaluated on the test set. We use AdamW \cite{loshchilov2018decoupled} optimizer with cosine learning rate scheduler and fixed number of warmup steps for optimization without gradient clipping.

\vspace{-0.3cm}
\section{Baselines}
\vspace{-0.2cm}

In this section, we provide additional details about the baseline methods. Very few approaches in the literature have attempted to geo-locate images on a scale of an entire world without any restrictions. In the last $5$ years, CNNs trained with large-scale datasets have significantly improved the planet-scale geo-localization performance. To the best of our knowledge, we are the first to introduce the effectiveness of fusion transformer architecture for this ill-posed problem. We compare our method with the following baselines:  
\begin{itemize}[leftmargin = 0.1in]
\item[$\bullet$] \textbf{Im$2$GPS} \cite{hays2008im2gps} is the first to attempt planet-scale geo-localization by using a simple retrieval approach to match a given query image based on a combination of different hand-crafted image descriptors to a reference dataset containing more than $6$M GPS-tagged images. This approach has later been improved \cite{hays2015large} by refining the search with multi-class support vector machines.
\item[$\bullet$] \textbf{PlaNet} \cite{weyand2016planet} is the first deep neural network trained for unconstrained planet-scale geo-localization. More specifically, PlaNet divides the earth in $26263$ geo-cells and trains an inception \cite{szegedy2015going} network with batch normalization \cite{ioffe2015batch} using $91$M geo-tagged images. PlaNet outperforms both versions of Im$2$GPS \cite{hays2008im2gps, hays2015large} by a substantial margin.
\item[$\bullet$] \textbf{[L]kNN} \cite{vo2017revisiting} proposes a retrieval-based geo-localization system which combines the Im$2$GPS and PlaNet by using features extracted by CNNs for nearest neighbour search. Though this method uses a $5$-times smaller training set than PlaNet, the retrieval-based approach requires a substantially larger inference time and disk space than classification approaches.   
\item[$\bullet$] \textbf{CPlaNet} \cite{seo2018cplanet} develops a combinatorial partitioning algorithm to generate a large number of fine-grained output classes by intersecting multiple coarse-grained partitionings of the earth. This technique allows creating small geo-cells while maintaining sufficient training examples per cell and hence improves the street- and city-level geolocational accuracy by a large margin.
\item[$\bullet$] \textbf{MvMF} \cite{izbicki2019exploiting} introduces the \textit{Mixture of von-Mises Fisher}
(MvMF) loss function for the classification layer that exploits the earth's spherical geometry and refines the geographical cell shapes in the partitioning.
\item[$\bullet$] \textbf{ISNs} \cite{muller2018geolocation} reduces the complexity of planet-scale geo-localization problem by leveraging contextual knowledge about environmental scenes. To deal with the huge diversity of images on earth's surface, this approach trains three different ResNet$101$ networks for \textit{natural}, \textit{urban}, and \textit{indoor} scenes, and achieves the current state-of-the-art performance on Im$2$GPS and Im$2$GPS$3$k. However, training different networks is cost-prohibitive and can not be generalized to a larger number of scenes. Our work addresses the limitations of ISNs by training a unified dual-branch transformer network in a multi-task framework and improves the state-of-the-art results by a significant margin.  

\end{itemize}

\vspace{-0.4cm}
\section{Evaluation Metrics}
\vspace{-0.2cm}
In a classification setup, we train \modelseg\ using cross-entropy loss which is closely associated with the classification accuracy. However, we evaluate \modelseg\ using geolocational accuracy. Hence, we empirically verify the strong correlation between these two metrics. We consider the Top-$N$ classification accuracy for $8$ different $N$ values ($1$, $5$, $10$, $50$, $100$, $200$, $300$, $500$) and geolocational accuracy ($a_r$) for $8$ different $r$ values ($1$, $25$, $100$, $200$, $400$, $750$, $1500$, $2500$) in similar intervals, and observe the correlation among them on different evaluation sets. The Pearson correlation coefficients between the two metrics for \modelseg\ are $0.978$, $0.984$, $0.985$ and $0.982$ on Im$2$GPS, Im$2$GPS$3$k, YFCC$4$k and YFCC$26$k, respectively. We also observe a similarly high correlation for the \modelrgb\ model. Figure $4$ in the main paper and Figure \ref{fig:metrics_supplementary} in supplementary material illustrates the strong linear correlation between the two metrics for \modelseg\ and \modelrgb\ on all $4$ evaluation sets.    

\begin{figure}[t!]
\centering
\hspace*{0cm}
\begin{center}
  \includegraphics[scale=0.105]{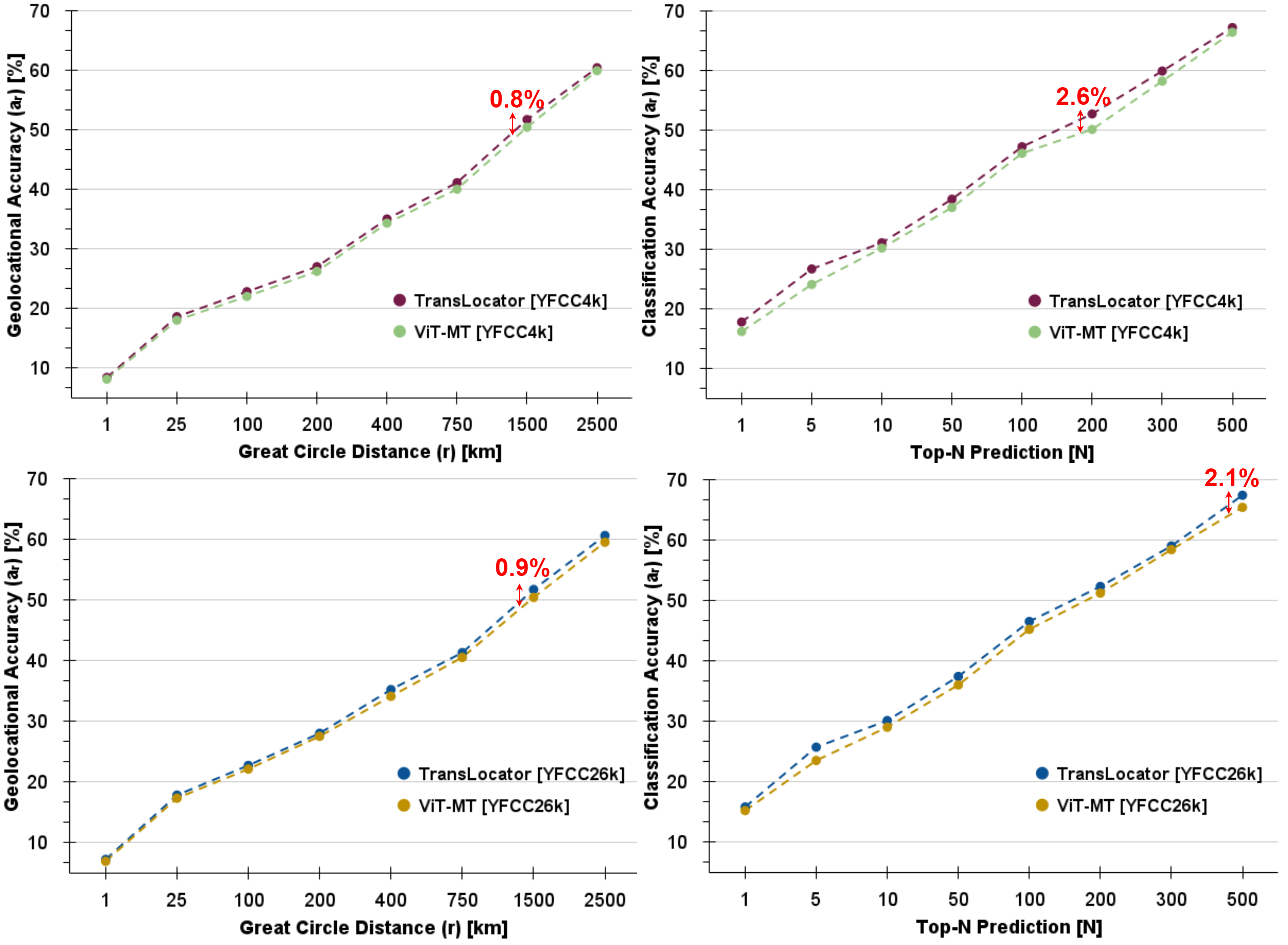}
\end{center}
\vspace{-0.5cm}
\caption{\textbf{Strong linear relationship between the geolocational and classification accuracy metrics.} The positive correlation enables us to treat geo-localization as a classification problem.}
\label{fig:metrics_supplementary}
\vspace{-0.4cm}
\end{figure}

\vspace{-0.2cm}
\section{Additional Quantitative Results}
\vspace{-0.2cm}
In this section, we investigate the contribution of different loss functions and provide a direct comparison of \modelseg\ with the ISNs \cite{muller2018geolocation}.

\vspace{-0.2cm}
\subsection{Ablation Study on Training Objective}

As discussed in section $3.4$ of the main paper, our overall training objective contains three losses for geo-cell prediction and one for scene recognition. Table \ref{tab:loss_ablation} shows how each loss function contributes to the performance of \modelseg\ on Im$2$GPS and Im$2$GPS$3$k datasets. We begin with training \modelseg\ separately on coarse, middle, and fine geo-cells. As the size of the geo-cells reduces, the geolocational accuracy with a smaller distance threshold typically improves. With the fine geo-cells, \modelseg\ gains $5$ - $8.3\%$ street-level accuracy than using the coarse cells. Combining all three different geo-cells helps the system learn geographical features at different scales, leading to a more discriminative classifier and improving the street-level accuracy by $0.2$ - $0.9\%$. The performance is further improved by another $0.7$ - $0.9\%$ after adding the scene information, which reduces the complexity of the data space by providing contextual knowledge about the surroundings.

\begin{table}[t!]
\centering 
\caption{\textbf{Ablation study on different losses of the training objective of \modelseg.} The finer geo-cells helps to improve the geolocational accuracy with smaller distance threshold. } 
\vspace{2mm}

\resizebox{0.775\columnwidth}{!}
{
\begin{tabular}{c | c c c c || c c c c c}
\hline
\multirow{3}{*}{\bf Dataset} & \multirow{3}{*}{\bf \centering $\mathcal{L}_{geo}^{corase}$} & \multirow{3}{*}{\bf \centering $\mathcal{L}_{geo}^{middle}$} & \multirow{3}{*}{\bf \centering $\mathcal{L}_{geo}^{fine}$} & \multirow{3}{*}{\bf \centering $\mathcal{L}_{scene}$} & \multicolumn{5}{c}{\bf Distance ($a_r$ [\%] @ km)} \\ 

& & & & & \multirow{1}{1.4 cm}{\tt \bf \centering Street} & \multirow{1}{1.4 cm}{\tt \bf \centering City} & \multirow{1}{1.4 cm}{\tt \bf \centering Region} & \multirow{1}{1.4 cm}{\tt \bf \centering Country} & \multirow{1}{1.4 cm}{\tt \bf \centering Continent} \\ 

& & & & & \bf $1$ km & \bf $25$ km & \bf $200$ km & \bf $750$ km & \bf $2500$ km \\

\hline
\hline

\multirow{5}{1.2 cm}{\tt \bf \centering Im$2$GPS\\ \cite{hays2015large}} & \cmark & \xmark & \xmark & \xmark & 9.8 & 30.9 & 54.4 & 72.5 & 84.7 \\
& \xmark & \cmark & \xmark & \xmark & 14.0 & 38.4 & 58.5 & 72.9 & \bf 86.7 \\
& \xmark & \xmark & \cmark & \xmark & 18.1 & 46.0 & 61.8 & 73.1 & 85.7 \\
& \cmark & \cmark & \cmark & \xmark & 19.0 & 47.2 & 62.7 & 73.5 & 85.7 \\
& \cmark & \cmark & \cmark & \cmark & \bf 19.9 & \bf 48.1 & \bf 64.6 & \bf 75.6 & \bf 86.7 \\

\hline
\hline

\multirow{5}{1.2 cm}{\tt \bf \centering Im$2$GPS\\$3$k\\ \cite{hays2015large}} & \cmark & \xmark & \xmark & \xmark & 5.9 & 22.7 & 40.7 & 54.9 & 77.0 \\
& \xmark & \cmark & \xmark & \xmark & 8.0 & 25.5 & 42.8 & 56.9 & 78.1 \\
& \xmark & \xmark & \cmark & \xmark & 10.9 & 29.8 & 45.0 & 56.8 & 78.1 \\
& \cmark & \cmark & \cmark & \xmark & 11.1 & 30.2 & 45.0 & 56.8 & 78.1 \\
& \cmark & \cmark & \cmark & \cmark & \bf 11.8 & \bf 31.1 & \bf 46.7 & \bf 58.9 & \bf 80.1 \\
\hline 
\hline

\end{tabular}}

\label{tab:loss_ablation}
\vspace{-0.3cm}
\end{table}

\begin{table}[h]
\centering 

\caption{\textbf{Comparison of unified and separate systems for different scene kinds.} Using separate systems is not only cost-prohibitive but also does not utilize semantic similarities across different scene kinds.}
\vspace{2mm}
\resizebox{0.9999\textwidth}{!}
{
\begin{tabular}{l | c | l || c  c  c || c c c || c c c}
\hline
\multirow{3}{*}{\bf Dataset} & \multirow{3}{*}{\bf Method} & \multirow{3}{1.4 cm}{\bf Train Set \{Scene\}} &  \multicolumn{9}{c}{\bf \centering Evaluation Set \{Scene\} (\bf Distance ($a_r$ [\%] @ km)} \\ 

& & & \multicolumn{3}{c||}{\bf Natural} & \multicolumn{3}{c||}{\bf Urban} & \multicolumn{3}{c}{\bf Indoor} \\ 

& & & \multirow{1}{1.2cm}{\bf \centering $1$ km} & \multirow{1}{1.2cm}{\bf \centering $200$ km} & \multirow{1}{1.2cm}{\bf \centering $2500$ km} & \multirow{1}{1.2cm}{\bf \centering $1$ km} & \multirow{1}{1.2cm}{\bf \centering $200$ km} & \multirow{1}{1.2cm}{\bf \centering $2500$ km} & \multirow{1}{1.2cm}{\bf \centering $1$ km} & \multirow{1}{1.2cm}{\bf \centering $200$ km} & \multirow{1}{1.2cm}{\bf \centering $2500$ km}  \\ 

\hline
\hline

\multirow{7}{1.4cm}{\tt \centering \bf Im$2$GPS\\ \cite{hays2008im2gps}} & \; \multirow{3}{*}{ISNs (M, f, S$_3$) \cite{muller2018geolocation}} \; & Natural & 2.5 & 48.8 & 71.3 & $-$ & $-$ & $-$ & $-$ & $-$ & $-$ \\ 

& & Urban & $-$ & $-$ & $-$ & 22.6 & 56.5 & 89.9 & $-$ & $-$ & $-$\\ 

& & Indoor & $-$ & $-$ & $-$ & $-$ & $-$ & $-$ & 15.8 & 31.6 & 57.9 \\

\cdashline{2-12}

& \multirow{3}{2 cm}{\centering TransLocator \\ w/o scene} & Natural & 3.8 & 54.9 & 77.1 & $-$ & $-$ & $-$ & $-$ & $-$ & $-$ \\ 

& & Urban & $-$ & $-$ & $-$ & 24.8 & 65.0 & 88.2 & $-$ & $-$ & $-$ \\ 

& & Indoor & $-$ & $-$ & $-$ & $-$ & $-$ & $-$ & 20.4 & 55.2 & 79.0 \\

\cdashline{2-12}


& \modelseg & All & 5.0 & 60.8 & 83.8 & 27.5 & 72.9 & 86.9 & 26.3 & 61.4 & 94.7 \\

\hline
\hline

\multirow{7}{1.4cm}{\tt \centering \bf Im$2$GPS\\$3$k\\ \cite{hays2015large}} & \; \multirow{3}{*}{ISNs (M, f, S$_3$) \cite{muller2018geolocation}} \; & Natural & 3.2 & 31.8 & 63.1 & $-$ & $-$ & $-$ & $-$ & $-$ & $-$ \\ 

& & Urban & $-$ & $-$ & $-$ & 14.1 & 44.8 & 72.7 & $-$ & $-$ & $-$\\ 

& & Indoor & $-$ & $-$ & $-$ & $-$ & $-$ & $-$ & 9.2 & 17.8 & 48.4 \\

\cdashline{2-12}

& \multirow{3}{2 cm}{\centering TransLocator \\ w/o scene} & Natural & 4.0 & 36.5 & 70.8 & $-$ & $-$ & $-$ & $-$ & $-$ & $-$\\ 
& & Urban & $-$ & $-$ & $-$ & 14.4 & 45.9 & 80.2 & $-$ & $-$ & $-$ \\ 

& & Indoor & $-$ & $-$ & $-$ & $-$ & $-$ & $-$ & 10.4 & 28.2 & 68.5 \\

\cdashline{2-12}


& \modelseg & All & 4.7 & 42.6 & 75.2 & 15.0 & 45.9 & 83.3 & 13.0 & 32.7 & 78.2 \\
\hline
\hline

\end{tabular}}

\label{tab:results_ablation_supplementary}
\vspace{-0.3cm}
\end{table}

\vspace{-0.2cm}
\subsection{Differences from ISNs}
\vspace{-0.2cm}

As the Individual Scene Networks (ISNs) \cite{muller2018geolocation} is the first method that utilizes scene information for geo-localization, we present clear differences of our method from ISNs. First, ISNs train three separate ResNet$101$ networks for natural, urban and indoor scenes. In contrast, \modelseg\ uses a unified dual-branch transformer backbone for all scenes. Using three different networks is cost-prohibitive and restricts the system from sharing the learned features across different scene kinds that likely have higher-order semantic similarities. To directly comprehend the effectiveness of a unified network, we train \modelseg\ without its scene recognition head separately on natural, urban and indoor images. As shown in Table \ref{tab:results_ablation_supplementary}, the single network achieves better performance than separate networks for all three scene kinds. Table \ref{tab:results_ablation_supplementary} also exhibits the effectiveness of dual-branch transformer backbone than ResNet for geo-localization. Moreover, unlike ISNs, the segmentation branch of \modelseg\ helps produce better qualitative performance under challenging real-world appearance variation.  

\vspace{-0.2cm}
\section{Additional Qualitative Results}
\begin{figure}[hb]
\centering
\hspace*{0cm}
\begin{center}
  \includegraphics[scale=0.136]{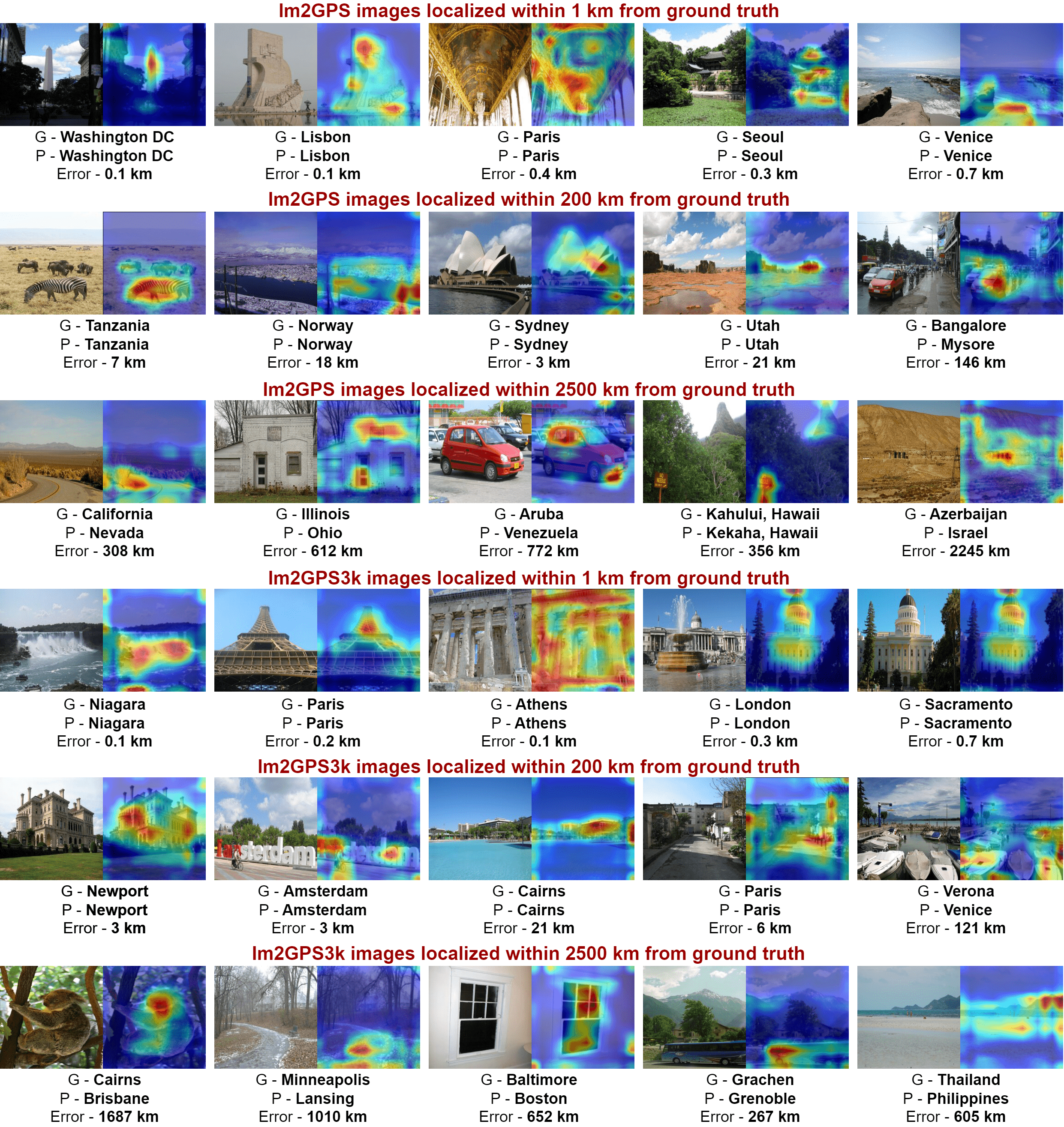}
\end{center}
\vspace{-0.5cm}
  \caption{\textbf{Example images from Im$2$GPS and Im$2$GPS$3$k dataset localized within three different distance threshold by \modelseg.} The corresponding Grad-CAM \cite{selvaraju2017grad} activation maps highlights the most important pixels used for prediction.}
  \label{fig:examples_supplementary}
\vspace{-0.4cm}
\end{figure}

\vspace{-0.2cm}
In this section, we visualize a few example images from the Im$2$GPS and Im$2$GPS$3$k datasets localized within $1$ km, $200$ km, and $2500$ km from ground truth locations by \modelseg\ in Figure \ref{fig:examples_supplementary}. The corresponding Grad-CAM \cite{selvaraju2017grad} activation maps highlight the necessary pixels used for the decision. Famous landmarks and tourist attractions like the Washington Monument, Eiffel Tower, Niagara Falls, and Trafalgar Square are correctly localized. Moreover, \modelseg\ often yields surprisingly accurate results for images with more subtle geographical cues, like a sea-beach in Venice or a uniquely-shaped building in Seoul. Minor errors like $100-200$ meters occur due to fine cells' size and can further be removed by using a retrieval extension. More difficult samples, like a forest in Tanzania and a desert in Utah, are localized within a range of $10$-$20$ km, which can be attributed to the bigger geo-cells in those sparsely-populated areas. \modelseg\ can learn to recognize famous streets, buildings, water-bodies, plants, animals and yields surprisingly good results even in remote locations.

\vspace{-0.75cm}

\begin{figure}[t!]
\centering
\hspace*{0cm}
\begin{center}
  \includegraphics[scale=0.112]{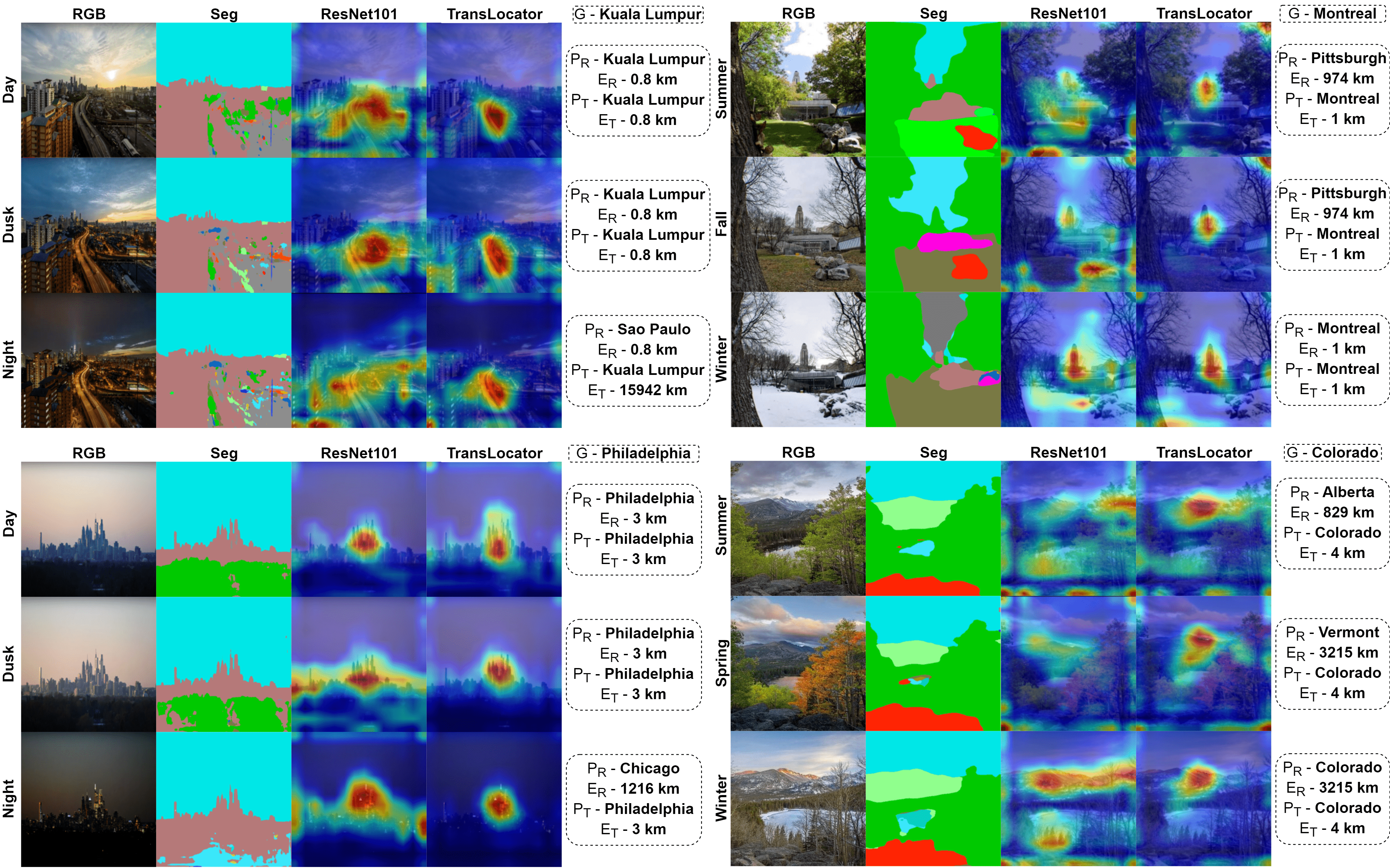}
\end{center}
\vspace{-0.5cm}
  \caption{\textbf{Qualitative comparison of \modelseg\ and ResNet$101$ on images with same location but under challenging appearance variations.} Unlike ResNet$101$, \modelseg\ attends to similar regions in each image and locates them correctly. \textit{G} denotes ground truth, \textit{$P_R$}, \textit{$E_R$}, \textit{$P_T$} and \textit{$E_T$} denotes predicted location and prediction error by ResNet$101$ and \modelseg, respectively. Best viewed when zoomed in and in color.}
  \label{fig:apperance_variation_supplementary}
\vspace{-0.6cm}
\end{figure}

Next, we illustrate a few more cases\footnote{Collected from the Internet under creative commons license.} of drastic appearance variation in the same location depending on the time of the day, weather, or season in Figure \ref{fig:apperance_variation_supplementary}. Though the RGB images experience extreme variation, the corresponding semantic segmentation maps remain unchanged. Thus, \modelseg\ can learn robust features and produce consistent activation maps across such radical appearance changes. In contrast, ResNet$101$ fails to recognize such variation. 

\end{document}